\documentclass[10pt,journal,compsoc]{IEEEtran}

\usepackage{cite}
\usepackage{amsmath,amssymb,amsfonts}
\usepackage{graphicx}
\usepackage{caption}
\usepackage{textcomp}
\usepackage{xcolor}
\def\BibTeX{{\rm B\kern-.05em{\sc i\kern-.025em b}\kern-.08em
    T\kern-.1667em\lower.7ex\hbox{E}\kern-.125emX}}
\usepackage{amsmath}
\usepackage{algorithm}
\usepackage[noend]{algpseudocode}
\usepackage{float} 
\usepackage{multirow}
\usepackage{mathrsfs}
\usepackage{CJK}
\usepackage{hyperref}

\usepackage{subcaption}
\usepackage{enumitem}
\usepackage{array}
\usepackage{colortbl}
\usepackage{tabularx}
\usepackage{ragged2e}
\usepackage{makecell}
\usepackage{lipsum} 

\newcommand{\comment}[1]{}

\begin{document}
%
\title{Graph Neural Networks for Tabular Data Learning: A Survey with Taxonomy \& Directions
}
%
%
%
%

\author{
        Cheng-Te Li~\IEEEmembership{Member,~IEEE}, 
        Yu-Che Tsai,
        Chih-Yao Chen,
        Jay Chiehen Liao
\IEEEcompsocitemizethanks{
\IEEEcompsocthanksitem Cheng-Te Li, Department of Computer Science and Information Engineering, National Cheng Kung University, Tainan, Taiwan. \\Email: chengte@ncku.edu.tw
\IEEEcompsocthanksitem Yu-Che Tsai, Department of Computer Science and Information Engineering, National Taiwan University, Taipei, Taiwan. \\Email: roytsai27@gmail.com
\IEEEcompsocthanksitem Chih-Yao Chen, University of North Carolina at Chapel Hill, NC, USA. Email: cychen@cs.unc.edu
\IEEEcompsocthanksitem Jay Chiehen Liao, Department of Computer Science and Information Engineering, National Cheng Kung University, Tainan, Taiwan. \\Email: jay.chiehen@gmail.com
}
\thanks{Github page: \url{https://github.com/Roytsai27/awesome-GNN4TDL}\\
Version date: January 1, 2024
}
}

\IEEEtitleabstractindextext{%
\begin{abstract}
In this survey, we dive into Tabular Data Learning (TDL) using Graph Neural Networks (GNNs), a domain where deep learning-based approaches have increasingly shown superior performance in both classification and regression tasks compared to traditional methods. The survey highlights a critical gap in deep neural TDL methods: the underrepresentation of latent correlations among data instances and feature values. GNNs, with their innate capability to model intricate relationships and interactions between diverse elements of tabular data, have garnered significant interest and application across various TDL domains.
Our survey provides a systematic review of the methods involved in designing and implementing GNNs for TDL (GNN4TDL). It encompasses a detailed investigation into the foundational aspects and an overview of GNN-based TDL methods, offering insights into their evolving landscape. We present a comprehensive taxonomy focused on constructing graph structures and representation learning within GNN-based TDL methods. In addition, the survey examines various training plans, emphasizing the integration of auxiliary tasks to enhance the effectiveness of instance representations.
A critical part of our discussion is dedicated to the practical application of GNNs across a spectrum of GNN4TDL scenarios, demonstrating their versatility and impact. Lastly, we discuss the limitations and propose future research directions, aiming to spur advancements in GNN4TDL. This survey serves as a resource for researchers and practitioners, offering a thorough understanding of GNNs' role in revolutionizing TDL and pointing towards future innovations in this promising area.
\end{abstract}

\begin{IEEEkeywords}
graph neural networks, tabular data learning, graph representation learning, graph structure learning, survey paper
\end{IEEEkeywords}}

\maketitle


\IEEEdisplaynontitleabstractindextext

%
\IEEEpeerreviewmaketitle

\ifCLASSOPTIONcompsoc
\section{Introduction}
\label{sec-intro}

The deep learning-based approaches to Tabular Data Learning (TDL), e.g., classification and regression, have exhibited promising performance in recent years~\cite{dnntbl21}. However, despite the great ability to learn effective feature representations from raw tabular records, the deep neural TDL provides weak modeling on the latent correlation among data instances and feature values. The prediction performance of TDL has been shown to improve by modeling high-order instance-feature relationships~\cite{you2020grape}, high-order feature interactions~\cite{fignn19}, and multi-relational correlation between data instances~\cite{tabgnn21}. As a natural antidote to model relations and interactions between different data entities, graph neural networks (GNNs) have recently received tremendous attention~\cite{gnn21tnnls}. By properly constructing the graph structures from the input tabular data, GNNs can learn latent correlations between data elements and generate effective feature representations for prediction tasks. Inspired by the success of GNNs on natural language processing~\cite{gnn4nlp} and recommender systems~\cite{gnnrec22}, there is also an increasing trend towards developing \textbf{Graph Neural Networks for Tabular Data Learning} (GNN4TDL).

Currently, a number of early research efforts have attempted to apply existing GNN methods for tabular data learning, e.g.,~\cite{lunar22,caregnn20,lstmgnn21,hsgnn20,xfraud21}. Some very recent studies~\cite{you2020grape,fignn19,tabgnn21,bgnn21,sublime22,slaps21} also have started to explore TDL-specific GNNs. These studies almost span all TDL topics and applications, setting off a wave of research enthusiasm in this field. Some essential questions also arise with this research progress. (a) \textit{What are the differences between GNN-based TDL and conventional TDL?} (b) \textit{What are the proper ways to construct graph structures under different TDL scenarios and tasks?} (c) \textit{What is the principle behind GNN-based tabular data representation learning?} (d) \textit{What are the TDL tasks and application domains that can benefit from GNNs?} (e) \textit{What are the limitations of current research and potential opportunities for future research?} Although encouraging results are reported in recent studies of GNN4TDL, these questions are not systematically investigated or even neglected. There is an urgent need for this GNN4TDL survey to reveal the answers to these questions in order to promote this line of research further.

\begin{table*}[!t]
\centering
\caption{A comparison between existing surveys on graph neural networks-based application domains. \textit{TDP}: tabular data prediction; \textit{GRL}: graph representation learning; \textit{GSL}: graph structure learning; \textit{SSL}: self-supervised learning; \textit{TS}: training strategy; \textit{AT}: auxiliary task; 
App: Applications.}
\label{tab:relatedsurv}
\begin{tabular}{l|c|c|c|c|c|c|c|c|c|c}
\hline
 & \textbf{Year} & \textbf{Domain} & \textbf{Data} & \textbf{TDP} & \textbf{GRL} & \textbf{GSL} & \textbf{SSL} & \textbf{TS} & \textbf{AT} & \textbf{App} \\ \hline
Wu et al.~\cite{gnn21tnnls} & 2021 & Graph Machine Learning & Graph &  & \checkmark &  &  &  &  & \checkmark \\ \hline
Zhang et al.~\cite{dlctr21} & 2021 & CTR Prediction & Tabular & \checkmark &  &  &  &  &  & \checkmark  \\ \hline
Wang et al.~\cite{glrs21} & 2021 & Recommender Systems & User-Item &  & \checkmark &  &  &  &  & \checkmark  \\ \hline
Borisov et al.~\cite{dnntbl21} & 2022 & Tabular Data Learning & Tabular & \checkmark &  &  & \checkmark &  &  & \checkmark \\ \hline
Wu et al.~\cite{gnnrec22} & 2022 & Recommender Systems & User-Item &  & \checkmark &  &  &  &  & \checkmark  \\ \hline
Zhu et al.~\cite{gsl22ijcai} & 2022 & Graph Machine Learning & Graph &  & \checkmark & \checkmark &  & \checkmark & \checkmark & \checkmark  \\ \hline
Zhang et al.~\cite{dlgsuv22} & 2022 & Graph Machine Learning & Graph &  & \checkmark &  &  & \checkmark & \checkmark & \checkmark  \\ \hline
Wu et al.~\cite{gnn4nlp} & 2023 & Natural Language Processing & Text &  & \checkmark & \checkmark &  &  &  & \checkmark  \\ \hline
Wu et al.~\cite{neurecsuv21} & 2023 & Recommender Systems & User-Item &  & \checkmark &  & \checkmark &  &  & \checkmark  \\ \hline
Ma et al.~\cite{gaddl21} & 2023 & Anomaly Detection & Graph &  & \checkmark &  &  & \checkmark &  & \checkmark  \\ \hline
Gao et al.~\cite{gnn4rsx21} & 2023 & Recommender Systems & User-Item &  & \checkmark &  &  &  &  & \checkmark  \\ \hline
Liu et al.~\cite{gssl22} & 2023 & Graph Machine Learning & Graph &  & \checkmark &  & \checkmark & \checkmark & \checkmark & \checkmark  \\ \hline
Dong et al.~\cite{gnniot22} & 2023 & Internet of Things & Sensor &  & \checkmark &  &  &  &  & \checkmark  \\ \hline
Jiao et al.~\cite{grlcv22} & 2023 & Computer Vision & Image &  & \checkmark &  &  &  &  & \checkmark  \\ \hline
Jin et al.~\cite{gnn4ts23} & 2023 & Time Series Analysis & Sensor &  & \checkmark & \checkmark &  &  &  & \checkmark  \\ \hline
This survey (GNN4TDL) & 2024 & Tabular Data Learning & Tabular & \checkmark & \checkmark & \checkmark & \checkmark & \checkmark & \checkmark & \checkmark  \\ \hline
\end{tabular}%
\end{table*}

We believe this GNN4TDL survey will be placed at a high value because of the high demand and low support on this topic. (a) \textit{High Demand}: since tabular data is ubiquitous in many domains and applications, and people have gradually shifted their focus to model the relationships between data instances and their correlation with feature values, we believe graph neural networks for tabular data learning shall have not only high research impact but also practical values. It shall be able to receive attention from both academia and industry. (b) \textit{Low Support}: Our GNN4TDL falls in a niche yet crucial area largely overlooked in previous surveys, according to a summary of comparisons in Table~\ref{tab:relatedsurv}. Unlike other works that concentrate on broader GNN applications across various domains and data types, this survey not only highlights the potential of GNNs in tabular data prediction, representation learning, and graph structure learning, but also pioneers in discussing self-supervised learning, various training strategies, and auxiliary tasks specifically in GNN4TDL. 

This survey paper presents an in-depth exploration of applying GNNs in tabular data learning. It first establishes the fundamental problem statement and introduces various graph types used to represent tabular data. The survey is structured around a detailed GNN-based learning pipeline, encompassing phases include \textit{Graph Formulation}, where tabular elements are converted into graph nodes; \textit{Graph Construction}, focusing on establishing connections within these elements; \textit{Representation Learning}, highlighting how GNNs process these structures to learn data instance features; and \textit{Training Plans}, discussing the integration of auxiliary tasks and training strategies for enhanced predictive outcomes. Beyond the review of GNN4TDL techniques, the survey further illustrates GNN applications in diverse fields, such as fraud detection and precision medicine, alongside a critical discussion on the current research limitations and future directions in the field of GNN4TDL.

We summarize the contributions of this survey as below. 
\begin{itemize}[leftmargin=*]
\item We provide a broad picture of the current development of graph neural networks for tabular data learning. A timely and comprehensive literature review is presented to help readers quickly grasp the basic concepts and step into this research field.
\item We organize existing arts of applying GNNs to tabular data learning. Particularly, we dive into why and how GNNs can better model tabular data and demystify the performance gains of tabular data classification and regression brought by GNNs. In practice, we highlight the basic guidelines for constructing various graph structures for tabular data modeling.
\item We demonstrate how GNN can be utilized in many tabular data application domains, such as fraud detection, precision medicine, click-through rate prediction, and handling missing data. We also provide academia and industry with an insightful discussion of the limitations in current research and future research directions on GNN4TDL.
\end{itemize}

We organize this paper as follows. Section~\ref{sec-pre} defines the related concepts used in the remaining sections. Section~\ref{sec-categorize} describes the framework of GNN4TDL and provides categorization from multiple perspectives. Section~\ref{sec-methods} systematically reviews existing GNN4TDL methods based on our categorization.
Section~\ref{sec-app} surveys the real-world applications of GNN4TDL in various domains.
Section~\ref{sec-directions} discusses the remaining challenges and possible future directions. Section~\ref{sec-conclude} concludes this survey in the end.

\section{Preliminaries}
\label{sec-pre}
This section covers the common and necessary elements of GNN4TDL. We first provide the general problem statement of tabular data learning (Sec.~\ref{subsec-tdl}). Then, we give the definitions and notations of various graphs commonly adopted to depict tabular data (Sec.~\ref{subsec-gnote}). We also describe the basic ideas of representation learning with GNNs (Sec.~\ref{subsec-gnn}), followed by different learning tasks (Sec.~\ref{subsec-task}) that can be used to design the training objectives in GNN4TDL. Besides, we further discuss why GNNs are required for TDL (Sec.~\ref{subsec-whygnntdl}).

\subsection{Tabular Data Learning}
\label{subsec-tdl}
We consider the supervised learning setting. Let $D=\{(\mathbf{x}_i,y_i)\}_{i=1}^{n}$ denote a tabular dataset with $n$ data instances, where $\mathbf{x}_i=(\mathbf{x}_i^{\text{(num)}}, \mathbf{x}_i^{\text{(cat)}})\in\mathcal{X}\subseteq\mathbb{R}^d$ represents numerical feature $x_{ij}^{\text{(num)}}$ and categorical features $x_{ij}^{\text{(cat)}}$, and $y_i\in\mathcal{Y}$ denotes the label of data instance $i$. The total number of features is denoted as $d$. The dataset can be split into three disjoint subsets, $D=D_{train} \cup D_{val} \cup D_{test}$, where $D_{train}$ is used for training, $D_{val}$ is used for early stopping and hyperparameter tuning, and $D_{test}$ is used for testing. The common tabular data learning tasks have three types: binary classification $\mathcal{Y}=\{0,1\}$, multi-class classification $\mathcal{Y}=\{1,2,...,C\}$, and regression $\mathcal{Y}=\mathbb{R}$, where $C$ is the number of class labels.

We assume that every input feature vector $\mathbf{x}_i$ in $D$ is sampled i.i.d. from a feature distribution $p_X$, and the labeled data pairs $(\mathbf{x}_i,y_i)$ in $D$ are drawn from a joint distribution $p_{X,Y}$. When some labeled samples from $p_{X,Y}$ are available, we can train a predictive model $f:\mathcal{X}\rightarrow\mathcal{Y}$ by minimizing the empirical supervised loss $\sum_{i=1}^{|D_{train}|}\mathcal{L}(f(\mathbf{x}_i),y_i)$, where $\mathcal{L}$ is a standard supervised loss function, e.g., cross-entropy (CE) for classification and mean squared error (MSE) for regression.

\subsection{Notations on Graphs}
\label{subsec-gnote}
We provide the definitions of different types of graphs with notations used in this paper. 

\textbf{Homogeneous Graph.} A homogeneous graph $\mathcal{G}=(\mathcal{V},\mathcal{E})$, where $\mathcal{V}=\{v_1,v_2,...,v_n\}$ ($|\mathcal{V}|=n$) is the set of nodes, $\mathcal{E}$ is the set of edges, and $\mathcal{E}\subseteq\mathcal{V}\times\mathcal{V}$. The set of neighbors of node $v_i$ is denoted as: $\mathcal{N}(v_i)=\{v_j\in\mathcal{V} | e_{i,j}\in\mathcal{E}\}$. The graph structure can also be depicted by an adjacency matrix $\mathbf{A}\in\mathbb{R}^{n\times n}$, where $\mathbf{A}_{i,j}=1$ indicates $e_{i,j}\in\mathcal{E}$ and $\mathbf{A}_{i,j}=0$ means $e_{i,j}\notin \mathcal{E}$.

\textbf{Attributed Graph.} An attributed graph contains features (e.g., attributes) associated with nodes or edges or both. In TDL, it is common to consider that only nodes have features and use $\mathbf{X}\in\mathbb{R}^{n\times d}$ to denote the node feature matrix. The attributed graph can be depicted by $\mathcal{G}=(\mathcal{V},\mathcal{E},\mathbf{X})$

\textbf{Heterogeneous Graph.} A heterogeneous graph $\mathcal{G}=(\mathcal{V},\mathcal{E})$ is a graph that contains different types of nodes or edges or both. Each node is assocaited with a mapping function $\phi_v(v_i)$ that maps $v_i\in\mathcal{V}$ into the corresponding type, i.e., $\phi_v(v_i):\mathcal{V}\rightarrow \mathcal{S}_v$, where $\mathcal{S}_v$ is the set of node types. Each edge is also assocaited with a mapping function $\phi_e(e_{i,j})$ that maps $e_{i,j}\in\mathcal{E}$ into the corresponding type, i.e., $\phi_e(e_{i,j}):\mathcal{E}\rightarrow \mathcal{S}_e$, where $\mathcal{S}_e$ is the set of edge types. $|\mathcal{S}_v|+|\mathcal{S}_e|>2$.

Note that bipartite, multi-partite, and multiplex graphs can be treated as three special types of heterogeneous graphs. For example, a bipartite graph is a heterogeneous graph with two types of nodes ($|\mathcal{S}_v|=2$) and a single type of edge ($|\mathcal{S}_e|=1$). A multi-partite graph is with multiple types of nodes ($|\mathcal{S}_v|>1$) and multiple types of edges ($|\mathcal{S}_e|=|\mathcal{S}_v|-1$). A multiplex graph has only one type of nodes ($|\mathcal{S}_v|=1$) and multiple types of edges ($|\mathcal{S}_e|>1$).

\subsection{Graph Neural Networks (GNNs)}
\label{subsec-gnn}
\textbf{Node Representation Learning in GNNs.} Given an attributed graph $\mathcal{G}=(\mathcal{V},\mathcal{E}, \mathbf{X})$, where $\mathbf{x}_i$ is the $d$-dimensional feature vector of node $v_i$, a GNN algorithm can learn to generate the node representation $\mathbf{h}_i$ for every node $v_i\in\mathcal{V}$ by implementing two functions. Suppose we are training a $K$-layer GNN, the node embedding at the $k$-th layer, i.e., $\mathbf{h}_i^{(k)}$, can be obtained via:
\begin{align*}
\mathbf{a}_i^{(k)} &= \text{aggregate}^{(k)}\left(\mathbf{h}_j^{(k-1)}: v_j\in\mathcal{N}(v_i)\right),\\
\mathbf{h}_i^{(k)} &= \text{combine}^{(k)}\left(\mathbf{h}_i^{(k-1)}, \mathbf{a}_i^{(k)}\right),
\end{align*}
where $\mathbf{h}_i^{(0)} = \mathbf{x}_i$, $\mathbf{h}_i=\mathbf{h}_i^{(K)}$, and $\text{aggregate}^{(k)}(\cdot)$ and $\text{combine}^{(k)}(\cdot)$ are the aggregation and combination functions at the $k$-th layer, respectively. The derived node representation $\mathbf{h}_i$ can be directly utilized for downstream tasks. Note that there are a variety of designs for aggregation and combination functions in different GNNs. One can refer to the general GNN survey paper~\cite{gnn21tnnls} to find how such two functions can be fulfilled. 

\textbf{Graph Representation Learning in GNNs.} Given an attributed graph $\mathcal{G}=(\mathcal{V},\mathcal{E}, \mathbf{X})$, along with the derived node representations $\mathbf{h}_i$ for every node $v_i\in\mathcal{V}$, we can generate the representation of the entire graph $\mathbf{h}_{\mathcal{G}}$ based on a \textit{readout} layer, given by:
\begin{equation*}
\mathbf{h}_{\mathcal{G}} = \mathcal{R}\left(\{\mathbf{h}_i | v_i\in\mathcal{V}\}\right),
\end{equation*}
where $\mathcal{R}(\cdot)$ is the readout function that maps the embeddings of nodes into a graph-level representation. The implementation of $\mathcal{R}(\cdot)$ can be a simple permutation-invariant function like summation or pooling methods (e.g., DiffPool~\cite{diffpool18} and TopKPool~\cite{topkpool19}).

\subsection{Downstream Tasks}
\label{subsec-task}
In this section, we discuss how to formulate tabular data representation learning as the problem of graph representation learning, and treat the tabular predictions as different graph-related downstream tasks. 
The common downstream tasks include node-level, edge-level, and graph-level variants. 
\begin{itemize}[leftmargin=*]
\item \textbf{Node-level Tasks} refer to node classification and node regression. The task aims at predicting the label $y_i\in\mathcal{Y}$ for every node $v_i\in\mathcal{V}$. Given the obtained node representation $\mathbf{h}_i$, by taking node classification as an example, the typical approach is to feed $\mathbf{h}_i$ into a multilayer perceptron (MLP) layer with Softmax function to produce prediction outcomes. The cross-entropy loss is typically adopted for model training.

\item \textbf{Link-level Tasks} refer to link prediction, edge classification, and edge regression. Link prediction, a binary classification task $y_{i,j}\in\mathcal{Y}=\{0,1\}$, predicts whether one node $v_i$ will connect to the other node $v_j$ given their corresponding embeddings $\mathbf{h}_i$ and $\mathbf{h}_j$. Edge classification aims to classify edges into multiple classes, and edge regression aims to regress an edge into a real value. For link prediction, a typical approach is to feed the vector concatenation of $\mathbf{h}_i$ and $\mathbf{h}_j$ into an MLP layer with the Softmax function to produce the outcome.

\item \textbf{Graph-level Tasks} refer to graph classification and graph regression, in which the input is graph representation $\mathbf{h}_{\mathcal{G}}$. Taking graph classification as an example, each graph $\mathcal{G}_i$ is associated with a target label $y_i\in\mathcal{Y}$. The goal is to train a model to predict its label. A typical approach is to feed $\mathbf{h}_{\mathcal{G}}$ into an MLP classifier with the Softmax function to generate the prediction results.
\end{itemize}

\subsection{Why are GNNs required for TDL?}
\label{subsec-whygnntdl}
Recent advances in tabular data learning have developed a rapidly growing branch: leveraging GNNs to obtain better instance representations to improve performance on downstream tasks. GNN-based tabular data learning methods have also achieved the state-of-the-art in various applications, such as click-through rate prediction~\cite{efignn22}, medical risk prediction with electronic health records~\cite{medgraph20}, anomaly detection~\cite{lunar22}, and missing data imputation~\cite{you2020grape}. We summarize why GNNs can benefit tabular data learning in the following five perspectives: (a) instance correlation, (b) feature interaction, (c) high-order connectivity, (d) supervision signal, and (e) inductive capability.

\textbf{Instance Correlation.}
Data instances can be correlated with each other in terms of their features. For example, users with similar profiles or online behaviors tend to have similar preferences for ads or items~\cite{rim21,dgenn21,grecplus22}. Patients with similar clinical data or symptoms have a higher potential to suffer from similar diseases~\cite{gct20,medgraph20,hcl22}. To better represent instances for downstream tasks, rather than solely employing each instance's self-features, it is crucial to model the correlation among instances. The key idea is to exploit such correlation to learn higher-quality feature representations of instances. Specifically, instances with similar downstream labels are close to one another, while those with different labels are pushed away from each other in the embedding space. With a proper construction of the graph structure that depicts the relationships between instances, GNN can be a good fit to let instances learn to represent each other. That said, we require GNNs to model instance correlation.

\textbf{Feature Interaction} refers to the effect of feature combination on the tabular prediction label. The positive contribution of one feature to the prediction task may depend on other features. Therefore, learning feature interactions can bring potential performance improvement. The conventional approach to obtain feature interactions is hand-crafted by enumerating some possible combinations of features, such as $\{\text{gender}=male\ \& \ \text{age}>20\}$ and $\{\text{gender}=female\ \& \ \text{age}>20\ \& \ \text{job}=lawyer\}$, which are second-order and third-order interactions, respectively. However, hand-crafting is time-consuming and requires domain knowledge. Deep neural network-based methods, such as Wide\&Deep~\cite{wnd16}, DeepFM~\cite{deepfm17}, Deep\&Cross~\cite{dnc17}, xDeepFM~\cite{xdeepfm18}, and Cross-GCN~\cite{crossgcn21}, can automatically learn feature interactions in an implicit fashion. Nevertheless, these methods merely combine the learned feature embeddings via simple concatenation. The \textit{structured correlation} among different features cannot be modeled. By considering features or instances as nodes in a graph, edges can depict the potential interactions among features or instances. GNN can naturally learn sophisticated feature interactions and produce a single embedding that captures the structured correlation among features.

\begin{figure}[!t]
  \centering
  \includegraphics[width=\linewidth]{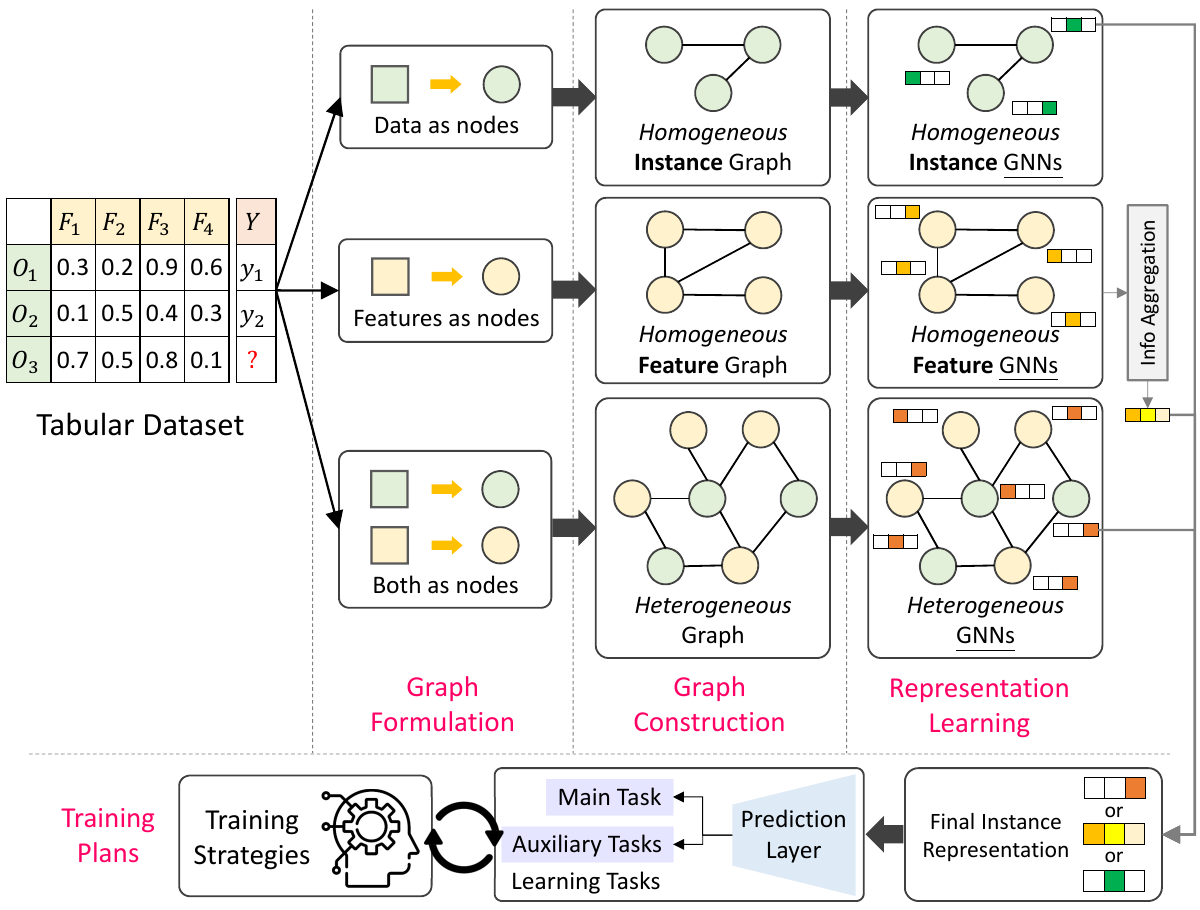} 
  \caption{The general pipeline of a GNN-based tabular data learning model. Given the tabular data as the model input, we first formulate the graph structure considering data instances and features, construct the graph structure, learn feature representations using GNNs, and train to generate the prediction outcomes, along with some auxiliary tasks.
  }
  \label{fig-flow}
\end{figure}

\textbf{High-order Connectivity.}
Tabular data learning aims at making predictions based on feature values. Instances possessing similar features tend to obtain the same prediction labels. In scenarios of utilizing deep neural networks in tabular data learning~\cite{dnntbl21}, likewise, deep models (e.g., VIME~\cite{vime20}, TabNet~\cite{tabgnn21}, SubTab~\cite{subtab21}, and SCARF~\cite{scarf22}) generate similar feature representations for instances with similar raw features in the learned embedding space such that they tend to be predicted as the same or close labels. These typical approaches implicitly realize such an idea based on the direct interactions between instances and their features in the training data. This utilization can be termed as the \textit{first-order connectivity}. To better learn feature representations of instances and to improve the prediction performance, the \textit{high-order connectivity} among instances~\cite{tabgnn21}, among features~\cite{fignn19}, and among instances and features~\cite{you2020grape} can be further considered. GNN-based methods can essentially model high-order interactions among data elements effectively. The utilization of high-order connectivity can be learned by the message passing and aggregation mechanisms in GNNs that receive embedding propagation from multi-hop neighbors in the graph.

\begin{figure*}[!t]
  \centering
  \includegraphics[width=\textwidth]{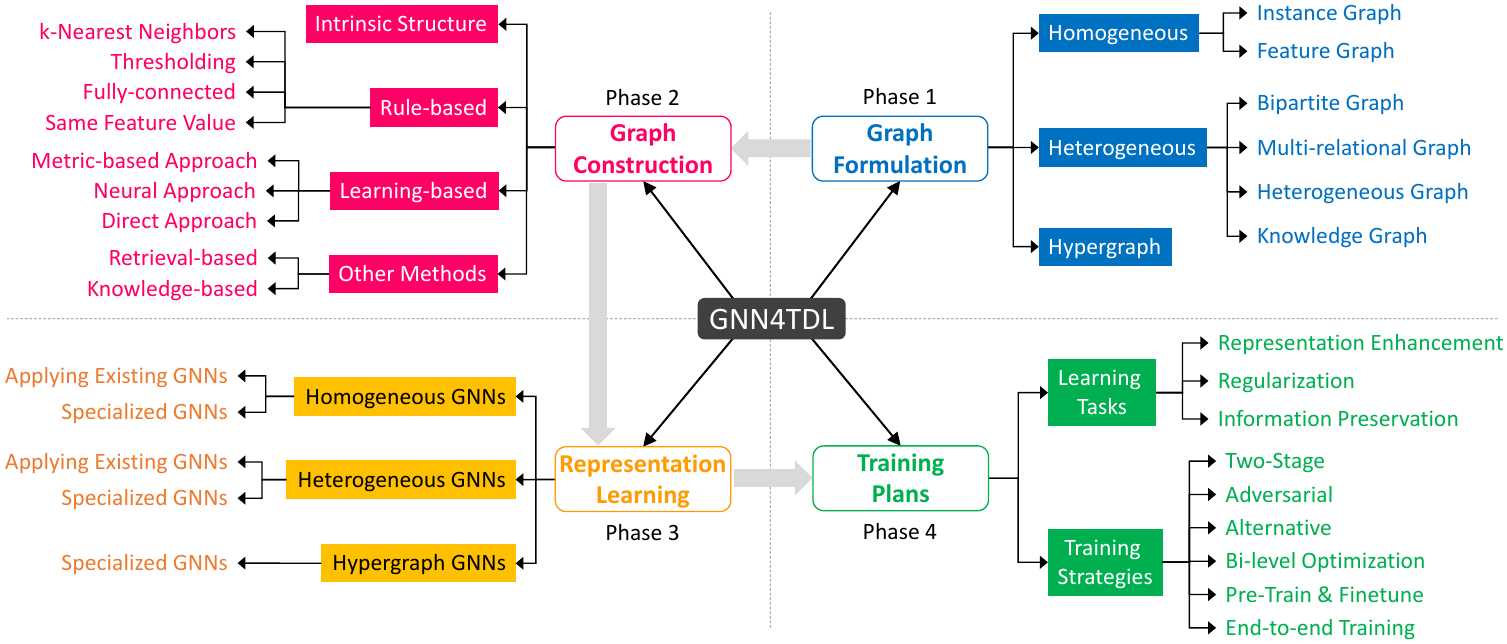} 
  \caption{The taxonomy systematically organizes GNNs for tabular data learning along four axes: graph formulation, graph construction, graph representation learning, and tabular training.
  }
  \label{fig-taxnomy}
\end{figure*}

\textbf{Supervision Signal.}
In real-world applications, such as fraud detection, medical prediction, and precision marketing, collecting a sufficient number of labeled tabular data is usually challenging. We face limited supervision signals when performing tabular data learning. One of the essential characteristics of graph neural networks is \textit{semi-supervised learning}, which can propagate the supervised information from the labeled instances to unlabeled ones via the graph structure so that better feature representations can be obtained and the supervision sparsity issue can be mitigated. Furthermore, recent advances in tabular data learning have attempted to incorporate \textit{self-supervised singals} into the representation learning process of tabular instances~\cite{vime20,subtab21,scarf22}. Developing auxiliary learning tasks on the features and having them trained together with the main task can further improve the performance of downstream tasks. A potential direction is to jointly design the self-supervised tasks based on both features and the graph structure and leverage the semi-supervised learning of GNNs.

\textbf{Inductive Capability.}
Graph neural networks exhibit the ability and \textit{inductive} power to generalize the message-passing mechanism to unseen nodes or even entirely new graphs~\cite{graphsage,gat18,graphsaint20}. A GNN model possesses inductive capability by adjusting its weights based on the supervised examples in the training data, and the model can be applied to new testing instances without further accessing the training set. Hence, when instances and features in a tabular dataset are represented as nodes, and their various interactions are treated as edges in a constructed graph, learning representations based on GNNs can make tabular data predictions capable of inductivity power. For example, the GNN-based tabular data learning will be able to deal with additional new features at the testing stage (i.e., feature extrapolation)~\cite{fate22}, produce the initialized embeddings for new data instances~\cite{gme21}, generalize to unseen tasks (not learned at the training stage)~\cite{metalink22}. Elements in tabular data will be endowed with inductive capability thanks to GNNs.

\section{Pipeline \& Categorization}
\label{sec-categorize}
In this section, we present the overarching pipeline of GNNs for tabular data learning, followed by a detailed categorization of each key phase, delineating how diverse methodologies fulfill these stages. Accompanying this categorization, we provide descriptions of select representative frameworks within each category. These studies exemplify the intricate interconnections among various phases or categories in the pipeline, highlighting their tight coupling and collaborative function in the overall GNN4TDL process. The detailed elaborations on every categorization are presented in Section~\ref{sec-methods}.

\textbf{Pipeline.} 
The general pipeline of GNN-based tabular data learning is provided in Figure~\ref{fig-flow}.
The pipeline begins with the \textit{Graph Formulation} phase, where the structure of the graph is defined using elements from the tabular dataset. This phase involves deciding which elements to use as nodes, with three common approaches: (1) representing data instances as nodes, (2) using features as nodes, or (3) a combination of both, forming different kinds of graph types. Following this, the \textit{Graph Construction} phase aims to create connections among these elements, transforming the tabular data into a graph structure. This structure is determined by the initial formulation, leading to either a homogeneous graph (e.g., instance graph or feature graph) or a heterogeneous graph (e.g., bipartite graph, multi-relational graph, or hypergraph).
Next, the \textit{Representation Learning} phase involves applying different types of GNNs based on the graph's nature. Various homogeneous instance GNNs, homogeneous feature GNNs, or heterogeneous GNNs are employed to learn feature representations of data instances. This phase is crucial as it determines how messages are propagated through the graph, modeling interactions among features and instances and influencing the quality of the learned embeddings. If the feature graph is used, an additional information aggregation layer is required to produce the final instance representation based on the learned embeddings of features.
Finally, the \textit{Training Plans} phase receives the final instance representations. In this phase, different learning tasks and training strategies are employed, including the use of auxiliary tasks alongside the main task. The outcome is then processed through a prediction layer to produce the final prediction outcomes. This comprehensive pipeline highlights the versatility of GNNs in handling various graph formulations and learning tasks, ultimately leading to effective tabular data learning and prediction.

\begin{table*}[!t]
\centering
\caption{List of representative GNN4TDL studies and their graph formulation settings. In column ``TDL Task'': Classification (Cla), Regression (Reg), Anomaly Detection (AD), Table Understanding (TU). Intrinsic (Sec.~\ref{ss-intri}), Rule (Sec.~\ref{ss-rule}), Learned (Sec.~\ref{ss-learn}). In column ``Graph Type'', Homogeneous (Homo) (Sec.~\ref{ss-homo}), Heterogeneous (Hete) (Sec.~\ref{ss-hete}), and Hypergraph (Sec.~\ref{ss-hyper}) graphs. 
}
\label{tab:representative}
\resizebox{\textwidth}{!}{%
\begin{tabular}{c|c|c|c|c|c|c|c}
\hline
 & Venue 'Year & Graph Type & Node & Edge & Node Initial Feature & Graph Task & TDL Task \\ \hline
FI-GNN~\cite{fignn19} & CIKM '19 & Homo & Feature & Rule & One-hot & Graph & Cla \\ \hline
GCT~\cite{gct20} & AAAI '20 & Hete & Instance, Feature value & Learned & Random & Graph & Cla \\ \hline
GNNDB~\cite{gnndb20} & ICLR '20 & Hete & Instance & Foreign key reference & Raw feat. & Node & Cla \\ \hline
IDGL~\cite{idgl20} & NeurIPS '20 & Homo & Instance & Learned & Raw feat. & Node & Cla \\ \hline
GRAPE~\cite{you2020grape} & NeurIPS '20 & Hete-Bipartite & Instance, Feature & Intrinsic & $\mathbf{1}$ (inst.), One-hot (feat.) & Node, Edge & Cla, Reg \\ \hline
HSGNN~\cite{hsgnn20} & BigData '20 & Hete & Instance, Feature value & Intrinsic & One-hot & Node & Cla \\ \hline
GME~\cite{gme21} & SIGIR '21 & Hete & Instance, Feature & Intrinsic & Random & Node & Cla \\ \hline
XFraud~\cite{xfraud21} & VLDB '21 & Hete & Instance, Feature value & Intrinsic & Raw feat. & Node & Cla \\ \hline
TabGNN~\cite{tabgnn21} & DLP-KDD '21 & Hete-Multiplex & Instance & Rule & Raw feat. & Node & Cla, Reg \\ \hline
FIVES~\cite{fives21} & KDD '21 & Homo & Feature & Search & One-hot & Graph & Cla \\ \hline
TabularNet~\cite{tabularnet21} & KDD'21 & Homo & Cell & WordNet & Pretrained BERT & Node & Cla \\ \hline
SLAPS~\cite{slaps21} & NeurIPS '21 & Homo & Instance & Learned & Raw feat. & Node & Cla \\ \hline
FATE~\cite{fate22} & NeurIPS '21 & Hete-Bipartite & Instance, Feature & Intrinsic & $\mathbf{0}$ (inst.), One-hot (feat.) & Node & Cla \\ \hline
DGM~\cite{dgm22tpami} & TPAMI '22 & Homo & Instance & Learned & Raw feat. & Node & Cla \\ \hline
LUNAR~\cite{lunar22} & AAAI '22 & Homo & Instance & Rule & Raw feat. & Node & AD \\ \hline
SUBLIME~\cite{sublime22} & WWW '22 & Homo & Instance & Learned & Raw feat. & Node & Cla \\ \hline
MetaLink~\cite{metalink22} & ICLR '22 & Hete-Bipartite & Instance, Task & Intrinsic & Raw Feat. (inst.), Random (task) & Node & Cla, Reg \\ \hline
PET~\cite{pet22} & NeurIPS '22 & Hete-Bipartite & Instance, Feature value & Intrinsic & $\mathbf{0}$ (inst.), Random (others) & Node & Cla \\ \hline
HCL~\cite{hcl22} & SDM '22 & Hypergraph & Feature value & Instance & One-hot & Edge & Cla \\ \hline
Table2Graph~\cite{table2graph} & IJCAI '22 & Homo & Feature & Learned & One-hot & Graph & Cla \\ \hline
T2G-Former~\cite{t2gformer23} & AAAI '23 & Homo & Feature & Learned & Tokenized Feature vector~\cite{fttrans21} & Graph & Cla, Reg \\ \hline
IGRM~\cite{igrm23aaai} & AAAI '23 & Hete-Bipartite & Instance, Feature & Intrinsic & $\mathbf{1}$ (inst.), One-hot (feat.) & Edge & Reg \\ \hline
EGG-GAE~\cite{egggae23} & AAAI '23 & Homo & Instance & Learned & Raw feat. & Node & Cls, Reg \\ \hline
DRSA-Net~\cite{drsanet23} & TKDE '23 & Homo & Feature & Learned & Feature value & Graph & Cla, Reg \\ \hline
HES-GSL~\cite{hesgsl23} & TNNLS '23 & Homo & Instance & Learned & Raw feat. & Node & Cla \\ \hline
GNN4MV~\cite{gnn4mv22} & TNNLS '23 & Homo & Instance & Rule & Raw feat. & Node & Cla \\ \hline
GraphFC~\cite{graphfc23} & CIKM '23 & Hete & Instance, Feature & Intrinsic & One-hot & Node & Cla, Reg \\ \hline
HyTrel~\cite{hytrel23} & NeurIPS '23 & Hypergraph & Cell & Intrinsic & Tokenized Feature vector~\cite{fttrans21} & Edge & TU \\ \hline
PALTO~\cite{ruiz2022tabular} & NeurIPS '23 & Homo & Feature & Other & Raw feat. & Graph & Cla \\ \hline
CCNS~\cite{ccns23neurips} & NeurIPS '23 & Homo & Instance & Rule & Raw feat. & Node & Cla \\ \hline
TabGSL~\cite{tabgsl23} & Arxiv '23 & Homo & Instance & Learned & Raw feat. & Node & Cla \\ \hline
RelBench~\cite{relbench}& Arxiv '23 & Hete & Entry & Primary-foreign key & Raw feat. & Node, Edge & Cla, Reg \\ \hline
\end{tabular}%
}
\end{table*}

\textbf{Categorization.}
The taxonomy of building graph neural networks for tabular data learning can be established according to the pipeline. We give the taxonomy in Figure~\ref{fig-taxnomy}. Below, we accordingly describe the categorizations in the taxonomy, in which some representative studies in each category are mentioned and summarized in Table~\ref{tab:representative}.
\begin{enumerate}[leftmargin=*]
\item The formulations of graphs from tabular data contain three main types: \textit{homogeneous graphs}, \textit{heterogeneous graphs}, and \textit{hypergraphs}. Based on data instances as nodes or features as nodes, in homogeneous graphs, we can formulate instance graphs (e.g.,~\cite{lstmgnn21,sublime22,tabgsl23}) and feature graphs (e.g.,~\cite{fignn19,table2graph,t2gformer23}), respectively. On the other hand, a heterogeneous graph can connect data instances to their corresponding features (and further to other metadata). The formulation of heterogeneous graphs can be either bipartite or multi-partite graphs~\cite{you2020grape,fate22,igrm23aaai,pet22}. One can also consider different feature values as various edge types, which depict different relationships between data instances, and thus formulate multiplex/multi-relational graphs~\cite{tabgnn21,pcgnn21,aognn22}. If a formulation allows data instances and all possible values of features to appear in a graph, one can construct heterogeneous graphs to represent the complicated information interdependency~\cite{hsgnn20,gnndb20,relbench,graphfc23}. As for the formulation of hypergraphs~\cite{hcl22,pet22,hytrel23}, tabular elements sharing the same attributes are linked by an edge. An edge in a hypergraph can join any number of tabular elements. For example, instances sharing the same feature values can be linked by an edge in a hypergraph.

\item Given a certain graph formulation, where nodes have already been determined, the second phase aims at constructing a graph by creating edge connections between nodes to substantiate that formulation. According to the criterion of edge creation, in general, there are four types of approaches, including \textbf{intrinsic structure}, \textbf{rule-based}, \textbf{learning-based}, and \textbf{other approaches}, where the first two types are widely adopted. The intuitive way to create links is to utilize the inherent relationships among tabular data elements, e.g., an instance contains feature values~\cite{you2020grape,fate22}, two instances share the same values of a specific feature~\cite{tabgnn21,wpn21}, a data tables is related to another via the primary-foreign key relationships~\cite{gnndb20,relbench}. To define the edges between data instances and/or features, the rule-based approach relies on some manually-specified heuristics, such as k-nearest neighbors~\cite{lunar22,lstmgnn21,gnn4mv22}, fully-connected structure~\cite{fignn19,ginn20,fingat21}, and thresholding~\cite{grin22,du2022graph}. The learning-based approach automatically generates edges between nodes. It can be divided into three sub-categories: 
The \textit{metric-based method} uses kernel functions to compute edge weights based on node similarities~\cite{dgm22tpami,egggae23}. The \textit{neural method} employs deep neural networks for adaptive graph construction~\cite{sublime22,tabgsl23,t2gformer23}. The \textit{direct method} views the adjacency matrix as learnable~\cite{lds19,allg22}.
Other approaches belong to either retrieval-based or knowledge-based. The retrieval-based approach resorts to 
either discover relevant and similar data instances to build the edges based on \textit{information retrieval} techniques~\cite{pet22}, or perform \textit{neural architecture search} to find a better graph topology for representation learning~\cite{fives21}. The knowledge-based approach requires domain experts to provide either the knowledge on the correlation between data instances~\cite{tabularnet21} or the knowledge graph that depicts the relationships among features~\cite{ruiz2022tabular}, so that the graph can be constructed in a fine-grained manner.

\item Once the graph that depicts tabular data is derived, no matter how data instances and their corresponding features are depicted via the graph structure, the next phase is to learn the final representation of each instance. Based on the type of the obtained graph, e.g., homogeneous or heterogeneous graphs, we can utilize \textit{homogeneous} GNN models (e.g., GCN~\cite{gcn17}, GraphSAGE~\cite{graphsage}, GAT~\cite{gat18}, and GIN~\cite{gin19}) and \textit{heterogeneous} GNN models (e.g.,  RGCN~\cite{rgcn18}, HGAT~\cite{hgat19}, and HGT~\cite{hgt20}) to produce the embedding of every instance. In addition to simply applying existing GNN models, some existing efforts have developed \textit{GNNs specialized} to better capture the various complicated interactions between instances and features (e.g., \cite{fignn19,lunar22,gct20,t2gformer23,gnn4mv22}). 

\item Designing a proper training plan based on the learned feature representation of instance is the last mile. The training plan can be discussed from two aspects, \textit{learning tasks} and \textit{training strategy}. While the main task is to predict the target label, a variety of supervision variants are developed to enhance the learning, and thus, different auxiliary tasks can be constructed. For example, leveraging the contrastive learning to better refine the graph structure learning~\cite{sublime22,tabgsl23}, introducing self-supervised learning with autoencoder to produce the denoised features~\cite{slaps21}, and imposing various graph regularizations to stabilize the graph learning and avoid overfitting~\cite{idgl20,allg22}. Since the data is inherently in the tabular form, additional learning tasks can preserve the properties in the input tabular data, such as global statistics of features~\cite{ginn20}, domain-knowledge preservation~\cite{medgraph20}, and spatial information encoding~\cite{tabularnet21}. 
A range of training strategies is employed to optimize GNN4TDL performance. Two-stage methods (e.g., \cite{sublime22}) sequentially learn graph structures and then train prediction models. Adversarial techniques (e.g., \cite{ginn20}) enhance the realism of feature reconstruction. Alternative approaches (e.g., \cite{gedi23}) dynamically adjust feature reconstruction weights for improved task relevance. Bi-level optimization (e.g., \cite{fate22}) concurrently tunes GCN parameters and graph generation. Pretrain-Finetune strategies (e.g., \cite{graphfc23}) leverage self-supervised learning for robust initial data understanding, followed by targeted finetuning, albeit with potential stage mismatches. End-to-end training (e.g., \cite{tabgnn21}) is the widest adopted strategy, offering streamlined learning-to-prediction processes, directly enhancing improved performance.

\end{enumerate}

\section{Methods}
\label{sec-methods}
In this section, by following the pipeline and taxonomy presented in Section~\ref{sec-categorize}, we describe the main ideas of various graph neural network-based methods for tabular data learning. We also highlight the weaknesses and discuss the advantages of each approach.

\subsection{Graph Formulation}
\label{subsec-graphform}

\subsubsection{Homogeneous Graphs}
\label{ss-homo}
Tabular data can be represented by a homogeneous graph so that the correlation between instances or the interactions between features can be captured. \textit{Instance graphs} are created to depict the relationships between data samples while \textit{feature graphs} can be constructed to model the associations between various features. 

\textbf{Instance Graphs.} 
Most of existing methods formulate tabular data samples and their correlation as instance graphs for graph neural networks (e.g., \cite{sublime22,slaps21,lstmgnn21,lunar22,idgl20,lds19,ginn20,tabgsl23,gnn4mv22,xiao2023graph,ccns23neurips,hesgsl23}). The construction of instance graphs relies on all features, and thus the graphs can been seen as modeling the \textit{global} relationships among data samples. That said, the instance graph is hard to capture the relationships between instances from multiple different aspects, i.e., \textit{local} persepctives in terms of feature subsets. In addition, we find that when creating instances, the numbers of features in the tabular datasets are usually small, e.g., $\leq 20$ in LSTM-GNN~\cite{lstmgnn21}, $\leq 64$ in SUBLIME~\cite{sublime22}, LUNAR~\cite{lunar22}, SLAPS~\cite{slaps21} and IDGL~\cite{idgl20}, $\leq 36$ in GINN~\cite{ginn20}, and $\leq 40$ in TabGSL~\cite{tabgsl23}. The potential reason is three-fold. The first is simplicity and interpretability. With a limited number of features, the relationships between nodes in the graph, in which each feature can be represented as an attribute of a node, can be more easily understood. The second is dimensionality issue. High-dimensional data can lead to overly complex graphs, which can be difficult to analyze and can lead to overfitting or the ``curse of dimensionality'' in graph structure learning~\cite{drgsl15}. The third is avoiding noise: In datasets with many features, not all features are necessarily meaningful or beneficial for the task at hand~\cite{wang2019learning}. Creating the instance graph with a large number of features can introduce noise that hinders the performance of graph neural networks.

\textbf{Feature Graphs.}
Modeling the interactions between features of a data instance as a homogeneous graph, in which nodes are features and edges are their correlation, is also widely adopted in the literature of tabular data learning
(e.g., \cite{fignn19,fives21,gcnint21,drsanet23,table2graph,t2gformer23,causalgnn23,ftgat23,ruiz2022tabular,ignnet23,he2023novel,ince23}). Most aim at learning high-order feature interactions based on the constructed feature graphs through stacking multi-layer GNNs and aggregate the feature embeddings to the instance representation for final predictions. A recent method, Casual-GNN~\cite{causalgnn23}, builds a causal feature graph to capture causal feature relationships based on causal discovery among field features. 
Two common ways are required to preprocess numerical and categorical features. One is transforming one-hot vectors of categorical features into low-dimensional dense vectors through field-wise embedding while directly feeding numerical features into the model (e.g., FI-GNN~\cite{fignn19}, GCN-INT~\cite{gcnint21}, Table2Graph~\cite{table2graph}, and Causal-GNN~\cite{causalgnn23}). The other relies on feature tokenizer~\cite{fttrans21}, such as T2G-Former~\cite{t2gformer23} and FT-GAT~\cite{ftgat23}. Factorization machines~\cite{fm10} can also be used to give initial embeddings of features~\cite{he2023novel}. Although the relationships between features can be learned in an automatic way, we may require significant domain knowledge to define the dependencies of some features, such as clinical variables in the medical~\cite{featdomain15} and financial domains~\cite{findomain21}. Incorrect or oversimplified dependencies could lead to poor performance. 

\subsubsection{Heterogeneous Graphs}
\label{ss-hete}
Heterogeneous graphs can depict complex and diverse relationships. They can depict instance-instance relationships, feature-feature relationships, as well as instance-feature relationships, providing a richer and more comprehensive representation of the tabular data. Considering nodes and edges can possess different information, the relevant methods can be divided into five categories in formulating a tabular dataset, including general heterogeneous graphs, bipartite graphs, multi-relational graphs, semantic graphs, multiplex graphs, and knowledge graphs.

\textbf{General Heterogeneous Graphs.}
Various information in a tabular data can be considered as different types of nodes to form a general heterogene graph. The basic idea is to take categorical feature values as nodes, along with the connections to nodes of data instances. Typical examples include product brands and item categories in click-through rate prediction~\cite{gme21}, diagnosis and treatment codes in electronic health records~\cite{gct20,hsgnn20}, importer companies and categories of goods in customs import declarations~\cite{graphfc23}, payment token and shipping address in credit card transactions~\cite{xfraud21}, and products and devices in online product reviews~\cite{cfath21}. The relational databases can be also coverted into a heterogeneous graph by treating rows as nodes, data tables as node types, and foreign key columns as edge types~\cite{gnndb20,relbench,zhang2023gfs}. The formulation into heterogeneous graphs can model the complex relationships between different types of entites. Nevertheless, not all categorical attributes can be utilized as nodes. A strategy to select categorical attributes as nodes is to conduct the \textit{homophilic tests}~\cite{baesens2015fraud}. While GNNs work well due to the homophily assumption~\cite{rossi2020proximity,wu2019simplifying}, i.e., most connections happen among nodes in the same class or with alike features~\cite{mcpherson2001birds}, attributes with strong homophilic effects are proper candidates to be used as nodes in the formulation of heterogeneous graphs.

\textbf{Bipartite Graphs.}
A bipartite graph is a type of graph where nodes are divided into two distinct groups, and connections (edges) only exist between nodes from different groups, not within a group. For a tabular dataset, one common formulation adopted by existing studies~\cite{featwalk19,you2020grape,medgraph20,dpgnn21,fate22,mgnn22} as a bipartite graph is to treat the instances (rows) as one set of nodes, the features (columns) as the other set of nodes, and feature values as edge weights. The other formulation is to take feature values as one set of nodes~\cite{pet22}, which is designed to enhance the interactions between features. Using a bipartite graph to depict tabular data offers several advantages. (a) By representing instances and features separately, the bipartite graph can maintain the original structure of the tabular data. (b) Bipartite graphs can handle diverse types of features (numerical, categorical, etc.) as different edge properties~\cite{you2020grape}. (c) Bipartite graphs offer a flexible way to incorporate new instances or features without requiring a restructuring of the entire dataset, and thus support incremental feature learning~\cite{fate22}. (d) Bipartite graphs can naturally tackle the issue of missing feature values by not creating the corresponding instance-feature links. (e) Link prediction (i.e., predicting whether a link should exist between instance and feature nodes) can be used to impute missing feature values~\cite{you2020grape}. (f) Bipartite graphs provide an efficient way to calculate the instance proximity, i.e., similarities between instances~\cite{featwalk19}. However, it is worth mentioning that transforming a large tabular dataset into a bipartite graph can be computationally expensive, and processing the resulting graph also requires specific techniques, such as incremental construction~\cite{bosaghzadeh2020incremental}, locality-sensitive hashing~\cite{fdknng20}, and FPGA-based Accelerator~\cite{fpgaacc23}.

\textbf{Multi-Relational Graphs.}
A tabular dataset can be depicted as a multi-relational graph that contain same-typed nodes and multi-typed edges (or termed ``relations''). We can connect data instances with one another according to the different kinds of relationships between instances. Consturcting the multi-relational graph is common in representing user-review tabular data~\cite{caregnn20,pcgnn21,aognn22,h2fd22}. Users are regarded as nodes, and edges depict various ways of user-user interactions, such as co-reviewing the same product~\cite{caregnn20,aognn22,h2fd22}, similar rating behaviors~\cite{caregnn20,pcgnn21,h2fd22}, one user trades to another~\cite{pcgnn21}, co-purchasing the same product~\cite{aognn22}, having the same symptom (symptom as the relation)~\cite{riognn22}, and having social relationship like family or workmate~\cite{pcgnn21}.
A special variant of multi-relational graph is \textit{multiplex graph}. While a multi-relational graph put all relations that connects instances in the same graph structure, a multiplex graph is a layered structure, in which all graph layers share the same set of nodes and every layer is a homogeneous graph whose edges are created by a certain relation. By considering sharing common or similar feature values as a kind of relation, TabGNN~\cite{tabgnn21}, AMG~\cite{amg20}, and GCondNet~\cite{gcondnet23} utilize multiplex graphs to formulate tabular datasets. An obvious benefit of multi-relational graph formulation is better Data Integration. Multi-relational graphs can be an effective way to integrate data from multiple sources or of different types. Different types of relationships from different sources can be represented with different types of edges, allowing for more comprehensive modeling of the data. Nevertheless, multi-relational graph formulation have the same challenge as general heterogeneous graph formulation -- requiring careful investigation on choosing feature or attribute values for relation creation.

\textbf{Knowledge Graphs.}
A knowledge graph (KG) offers powerful auxiliary information to model the relationships between features of a tabular dataset. In KG formulation of tabular data, each feature corresponds to a node in the graph, and there are also non-feature nodes. One can access or construct the relevant KG from external resources and domain knowledge, such as human protein map~\cite{luck2020reference} and DrugBank~\cite{wishart2018drugbank} for bio-medical tabular data. PLATO~\cite{ruiz2022tabular} incorporates auxiliary domain information structured as a knowledge graph to regularize a multilayer perceptron for tabular data prediction. PLATO had shown that when working with tabular datasets with extremely high-dimensional features and limited samples, KG can mitigate the overfitting issue~\cite{ruiz2022tabular}. The challenge of adopting KG lies in finding the correspondences between tabular features and KG nodes. By tackling the scarcity of context and the ambiguity and noisiness of the available conten, JenTab~\cite{jentab20} learns to match tabular data to a knowledge graph.

\subsubsection{Hypergraphs}
\label{ss-hyper}
A hypergraph is a generalization of a graph in which an edge (or a hyperedge) can connect more than two nodes. When applying this to a tabular dataset, each feature value can be considered as a node, and each instance (or row of the dataset) is used to create a hyperedge that connects the nodes corresponding to the feature values of that instance~\cite{hcl22,pet22,hytrel23,hypervis19}. This representation can offer a more accurate depiction of complex relationships among feature values across instances. By treating each instance as a hyperedge, we effectively encode the simultaneous co-occurrence of multiple feature values. This can be particularly beneficial in cases where interactions among feature values are of interest. In addition to utilize every instance to create a hyperedge, HYTREL~\cite{hytrel23} further leverages each feature column and the entire tabular dataset to be two additional types of hyperedges. As pointed out by HYTREL~\cite{hytrel23}, hypergraphs with three types of hyperedges can preserve four structural properties in tabular data: (1) capture the permutation invariance of tables, maintaining relationships regardless of row or column order; (2) highlight interdependencies within rows and structural similarities within columns; (3) high-order multilateral relations within cells, rows, or columns are effectively represented; (4) depict the hierarchical organization of information in tables from cell-level to table-level. There are still limitations in hypergarph formulation. The first is sparse data issue. In high-dimensional datasets with many unique feature values, the hypergraph can become very large and sparse, which could pose challenges in analysis and computation. The second is scalability. As the size of the dataset grows, so does the complexity and size of the hypergraph. Handling large hypergraphs can be computationally challenging.



\subsection{Graph Construction}
\label{subsec-graphcons}
Graph construction is a critical step when transforming tabular data into graph data, and the chosen methodology can greatly affect the quality of graph structure. This in turn can have a significant impact on downstream prediction tasks. The key consideration is selecting or learning a proper criteria for edge creation, which can significantly impact the sparsity and connectivity of the graph. Too many edges can make the graph overly complex and computationally intensive to work with, while too few can lead to a highly disconnected graph.

\subsubsection{Intrinsic Structure}
\label{ss-intri}
A tabular dataset typically consists of instances (rows) and features (columns). The intrinsic structure of a tabular dataset emerges from the relationships between these instances and features. It is intuitive and common to utilize the intrinsic structure of tabular data to construct the graph. The formulation of bipartite graphs directly employs the intrinsic structure to create edges between instance nodes and feature nodes. Typical bipartite graph formulations, including GRAPE~\cite{you2020grape}, MedGraph~\cite{medgraph20}, FATE~\cite{fate22}, PET~\cite{pet22}, GNNDP~\cite{dpgnn21}, MGNN~\cite{mgnn22}, and IGRM~\cite{igrm23aaai}, leverage this approach to construct the graph. In addition, the constructions of general heterogeneous graphs and multi-relational graphs also rely on the intrinsic structure of tabular data. If an instance possess a specific categorical feature value, which is used as a certain node type, we can create an edge between them. 

In general heterogeneous grapohs, for example, edges encapsulate the interactions between user nodes, product brands, and item categories in click-through rate prediction~\cite{gme21}. This approach is mirrored in electronic health records, where edges are created between patient nodes, and diagnosis and treatment codes~\cite{gct20,hsgnn20}. Customs import declarations are depicted as connections between declaration nodes, importer companies, and goods categories~\cite{graphfc23}. Similarly, in credit card transactions, edges signify the relationships between transaction nodes, payment tokens, and shipping addresses~\cite{xfraud21}. In real estate valuation, edges depict the relationships, including location, facility, or floorplan, between houses entities~\cite{lifeprice23tkde}. In predicting stock price movements, various lead-lag relationships between companies, such as historical prices and media event, can be represented as edges~\cite{magnn22pr}. For supply chain mining, the ownerships of enterprises and their owners, the transfers/transactions between enterprises and between consumers, and the social relations between the owners of enterprises, are used to build the graph~\cite{yang2021financial}. In online product reviews, edges connect user nodes with corresponding products and device nodes, encapsulating user interactions~\cite{cfath21}. Last, in the modeling of relational databases, while instances or entries are used as multi-typed nodes, the primary-foreign key relations between data tables are used to define edges~\cite{gnndb20,relbench}.

In multi-relational graphs, for example, two user instances can be connected with each other because both had ever purchased or reviewed that product, referring to user-user \textsf{purchase} or \textsf{review} relationships~\cite{caregnn20,aognn22,h2fd22}. One user may trade something to another, indicating the \textsf{trade} relationship between them~\cite{pcgnn21}. A patient has a common symptom with another, employing the specific \textsf{symptom} as the relationship~\cite{riognn22}. In financial domains, companies can have explicit stock relations (e.g., supplier-consumer relations), which reflect the potential influence between stocks~\cite{tgc19tois,sgrn22ijcai,kim2019hats}. Users can be connected by multiple relationships, including social connections, capital transactions, and device dependence, for loan fraud detection~\cite{xu2021towards}. Various relations like transaction, transfer
and social can be used to build the multiplex graph among users for loan default prediction~\cite{amg20}.

The intrinsic structure among data instances, feature columns, feature values, and data cells can be naturally used to constuct a hypergraph. 
In HCL~\cite{hcl22}, each data instance is viewed as a hyperedge, and each categorical feature value is viewed as a node. HyTrel~\cite{hytrel23} treats each cell as a node, and each data instance, each feature column, and the entire data table as various hyperedges, respectively. That said, each cell node is connected to three hyperedges. In PET~\cite{pet22}, each distinct feature value is considered as a node, and each data instance constructs a hyperedge.

Utilizing the intrinsic structure of tabular data for graph construction has its limitations, largely due to the assumption that relationships between instances and features are clearly discernible. However, this may not always be the case due to various data challenges. For example, data noise, such as erroneous entries or outliers, can obscure true relationships. Imagine a tabular dataset of sensor readings where occasional malfunctioning sensors contribute incorrect data -- these noisy entries could disrupt the accurate capture of relationships in a graph structure. Another challenge is missing values, which can leave gaps in the connections. For instance, in a retail dataset, if certain transactions do not record the customer's age or the purchased product category, these missing values would make it difficult to form complete instance-feature connections in the graph. Furthermore, the heterogeneity of features, i.e., different types and scales of features, can complicate the relationships. In a real estate dataset, features might range from categorical (e.g., house style) to continuous (e.g., house size), and from ordinal (e.g., condition rating) to binary (e.g., has pool or not). The different nature of these features could pose challenges in defining clear and meaningful connections for graph construction. Moreover, the intrinsic structure may not capture latent relationships, which are indirect or hidden relationships that are not explicitly represented in the data. For example, in a movie rating dataset, there might be a latent relationship between two users who haven't rated the same movies but have similar tastes inferred through other shared features or behaviors. These latent relationships can be crucial to uncover underlying patterns in the data. Hence, although the intrinsic structure provides a basis for graph construction, it is important to consider these potential issues and complement this approach with additional techniques for handling them effectively.

\begin{table}[!t]
\centering
\caption{Summary of rule-based methods for tabular graph construction. In the column of Tabular Domain, ``multi'' means multiple domains.}
\label{tab:rule-construct}
\resizebox{\columnwidth}{!}{%
\begin{tabular}{c|c|c|c|c}
\hline
 & Formulation & Similarity & Edge Criteria & Tabular Domain \\ \hline
LSTM-GNN~\cite{lstmgnn21} & Ho-Inst & TF-IDF & kNN & medical \\ \hline
LUNAR~\cite{lunar22} & Ho-Inst & Euclidean & kNN & multi \\ \hline
GNNBFD~\cite{xiao2023graph} & Ho-Inst & Euclidean & kNN & manufacture \\ \hline
MST-GRA~\cite{gramst22} & Ho-Inst & Heat Kernel & kNN & multi \\ \hline
GCIL~\cite{gimblearn21} & Ho-Inst & RBF Kernel & kNN & multi \\ \hline
SSGNet~\cite{ssgnet22} & Ho-Inst & Euclidean & kNN & medical \\ \hline
GNN4MV~\cite{gnn4mv22} & Ho-Inst & Gaussian Kernel & kNN & multi \\ \hline
GRIN~\cite{grin22} & Ho-Inst & Correntropy~\cite{liu2007correntropy} & kNN & time-series \\ \hline
GINN~\cite{ginn20} & Ho-Inst & Euclidean & Threshold & multi \\ \hline
GAEOD~\cite{du2022graph} & Ho-Inst & Cosine & Threshold & multi \\ \hline
GEDI~\cite{gedi23} & Ho-Inst & Cosine & Threshold & time-series \\ \hline
LatentG~\cite{koloski2023latent} & Ho-Inst & Cosine & Threshold & medical \\ \hline
CCNS~\cite{ccns23neurips} & Ho-Inst & Cross-class sim~\cite{ma2022is} & kNN & multi \\ \hline
GATE~\cite{gate20} & He & PMI & Threshold & medical \\ \hline
WPN~\cite{wpn21} & Ho-Inst & - & Same Feat Val & medical \\ \hline
TabGNN~\cite{tabgnn21} & He-MultiRel & - & Same Feat Val & multi \\ \hline
SGANM~\cite{sganm22} & Ho-Inst & - & Fully-connected & manufacture \\ \hline
IAGNN~\cite{iagnn21} & Ho-Inst & - & Fully-connected & manufacture \\ \hline
GCN-Int~\cite{gcnint21} & Ho-Feat & - & Fully-connected & CTR \\ \hline
FinGAT~\cite{fingat21} & Ho-Feat & - & Fully-connected & financial \\ \hline
Fi-GNN~\cite{fignn19} & Ho-Feat & - & Fully-connected & CTR \\ \hline
INCE~\cite{ince23} & Ho-Feat & - & Fully-connected & multi \\ \hline
IGNNet~\cite{ignnet23} & Ho-Feat & Pearson correlation & Fully-connected & multi \\ \hline
GatFM~\cite{he2023novel} & Ho-Feat & Inner Product & Threshold & CTR \\ \hline
\end{tabular}%
}
\end{table}

\subsubsection{Rule-based Methods}
\label{ss-rule}
Rule-based methods for constructing graphs from tabular data involve the use of predefined criteria or heuristics to determine the nodes and edges of the resulting graph. Typically, instances (rows) and features (columns) of the table serve as the potential nodes, while relationships, often determined by logical conditions or thresholds, form the edges. For instance, an edge might be created between two instance nodes if they share a certain categorical feature value or if the simiarlity between their feature vectors exceeds a specific threshold. The rules can also be based on domain knowledge, correlations, or statistical properties of the data. The advantage of rule-based methods is that they offer explicit control over the graph construction process, ensuring that the resultant structure aligns well with domain knowledge or specific analytical objectives. However, crafting effective rules requires a deep understanding of both the data and the desired outcomes, and a poorly defined set of rules might lead to unrepresentative or overly sparse graph structures.

The basic steps of rule-based method include utilizing a certain simiarlity measure between data and determining some criteria for edge creation. By identifying the strategy to have the similarity measure and determine the edge criteria, we summarize the rule-based methods in Table~\ref{tab:rule-construct}. We can see that while different similarity measures can be used, there are four mainstream criteria of edge creation, including kNN, thresholding, fully-connected, and same feature value. 
\begin{itemize}[leftmargin=*]
\item \textbf{k-Nearest Neighbors (kNN)} (e.g., \cite{lstmgnn21,lunar22,gnn4mv22,xiao2023graph,ccns23neurips}) connects a node to its $k$ closest nodes based on some similarity metric in the feature space. kNN primarily focuses on local structure, ensuring that nodes with similar characteristics are connected, and preserving the local nuances of the dataset. Hence, kNN is often useful for datasets where instances have clear clusters or groupings. For large datasets, finding the $k$ nearest neighbors can be computationally expensive. It is also sensitive to the choice of the distance metric and the value of $k$. An inappropriate distance measure might not capture true data similarities.

\item \textbf{Thresholding} (e.g., \cite{gate20,grin22,du2022graph,gedi23}) forms edges between nodes when their similarity/distance value exceeds or falls below a set threshold. By setting a threshold, one can reduce the number of edges to eliminate weaker or spurious connections, resulting in a sparser and potentially more manageable graph. However, the choice of threshold can be arbitrary and might not always reflect inherent data structures. Overly aggressive thresholding can result in loss of important relationships.

\item \textbf{Fully-connected} (e.g., \cite{sganm22,fingat21,gcnint21,iagnn21}) makes every node in the graph connected to every other node, creating a dense web of relationships. This method can maximize the potential information flow, and capture global properties of the dataset. However, this can lead to a very dense graph, which might be computationally expensive to process. The presence of all possible connections can dilute the importance of truly significant relationships.

\item \textbf{Same Feature Value.} (e.g., \cite{wpn21,tabgnn21}) Edges are drawn between nodes that share an identical value for a particular feature. The reason for the connections is clear and easily interpretable. It is effective for categorical or discrete data. However, this can lead to isolated clusters in the graph, especially if many unique feature values exist. This is not always effective for continuous features without discretization.
\end{itemize}

\begin{table}[!t]
\centering
\caption{Comparison of learning-based methods for tabular graph construction. All of these methods are for the construction of instance graphs.}
\label{tab:learn-construct}
\resizebox{\columnwidth}{!}{%
\begin{tabular}{c|c|c|c|c}
\hline
Method & Strategy & Initialization & Modeling & Training \\ \hline
IDGL~\cite{idgl20} & Metric & - & Cosine & End-to-end  \\ \hline
RGSL~\cite{rgsl22} & Metric & - & Inner product & End-to-end \\ \hline
DGM~\cite{dgm22tpami} & Metric & kNN & Euclidean/Hyperbolic & End-to-end \\ \hline
EGG-GAE~\cite{egggae23} & Metric & - & Gaussian Kernel & End-to-end \\ \hline
DRSA-Net~\cite{drsanet23} & Metric & - & Low-rank Approx. & End-to-end \\ \hline
HES-GSL~\cite{hesgsl23} & Metric & kNN & Cosine & End-to-end \\ \hline
kNNGNN~\cite{kang2021k} & Metric & kNN & Mahalanobis & End-to-end \\ \hline
SLAPS~\cite{slaps21} & Neural & kNN &  Multilayer perceptron & End-to-end  \\ \hline
SUBLIME~\cite{sublime22} & Neural & kNN & Contrastive Learner & Unsupervised \\ \hline
TabGSL~\cite{tabgsl23} & Neural & kNN & Contrastive Learner & End-to-end \\ \hline
T2G-Former~\cite{t2gformer23} & Neural & Threshold & Multilayer perceptron & End-to-end \\ \hline
LDS~\cite{lds19} & Direct & kNN & Free variables & Alternative  \\ \hline
ALLG~\cite{allg22} & Direct & kNN & Free variables & Unsupervised Active \\ \hline
Causal-GNN~\cite{causalgnn23} & Direct & Random & Free variables & End-to-end \\ \hline
Table2Graph~\cite{table2graph} & Direct & Random & Self Attention~\cite{sapos18} & End-to-end \\ \hline
\end{tabular}%
}
\end{table}

\subsubsection{Learning-based Methods}
\label{ss-learn}
Given a tabular dataset, learning-based methods seek to derive the optimal structure of a graph, aiming to capture underlying relationships among data instances effectively. Instead of relying solely on pre-defined rules or domain-specific knowledge, this approach focuses on automatically determining the relationships between instances based on patterns present in the data. We can categorize existing studies into the following three groups, and accordingly create a summary of comparison in Table~\ref{tab:learn-construct}.
\begin{itemize}[leftmargin=*]
\item \textbf{Metric-based Approach} (e.g., \cite{idgl20,rgsl22,dgm22tpami,egggae23,hesgsl23,drsanet23}) leverages kernel functions to compute the similarity between node features or embeddings, and use these similarities as edge weights. Built on the network homophily principle, which posits that similar nodes are more likely to connect~\cite{newman2018networks}, metric-based techniques reinforce intra-class connections, resulting in denser and more concise graph representations. A significant advantage is that they often allow end-to-end training due to the differentiable nature of kernel functions. However, selecting an appropriate kernel function and its parameters can be non-trivial and can greatly influence results. In addition, the assumption of network homophily might not always hold for every dataset.
\item \textbf{Neural Approach} (e.g., \cite{slaps21,sublime22,tabgsl23,t2gformer23}) employs sophisticated deep neural networks to determine edge weights based on node features and their representations. This results in a more dynamic and adaptive graph construction mechanism as it can adaptively model complex relationships, potentially capturing non-linearities and intricate patterns. However, they may overfit to noise in the data if not properly regularized or if the dataset is small.
\item \textbf{Direct Approach} (e.g., \cite{lds19,allg22,causalgnn23,table2graph}) view the adjacency matrix of the desired graph as a set of learnable parameters. Unlike the previous methods that depend on node representations to define edge connections, direct approaches offer greater adaptability in graph construction due to no reliance on node representations. However, this increased flexibility often comes at the cost of difficulty in efficiently learning the matrix parameters. Without reliance on explicit features or patterns, these methods might sometimes incorporate noise or irrelevant relationships.
\end{itemize}

\subsubsection{Other Methods}
\textbf{Retrieval-based.}
The retrieval-based graph construction method~\cite{pet22,fives21} specifically captures the relationships within tabular data. In PET~\cite{pet22}, for each data row (target), relevant neighboring rows are identified from a data pool. These rows then form a hypergraph where sets of feature columns become hyperedges and individual feature values become nodes. This structure uniquely captures relationships, such as how two rows might share a particular feature value (a node). One significant advantage is the nuanced representation of data: whereas traditional tabular data might just list values, the hypergraph visually and structurally links related data points, offering a richer context and potentially unveiling hidden patterns and connections.
Feature Interaction Via Edge Search (FIVES)~\cite{fives21} automatically generates high-order interactive features from tabular data. It formulates the interactive feature generation as an edge search task on a feature graph. Utilizing a dedicated GNN and an adjacency tensor, FIVES inductively searches for optimal higher-order features based on interactions between existing features. This method transforms feature space traversal into an efficient GNN training process.

\textbf{Knowledge-based.}
For tabular data in domains like medicine or genomic, rows might represent patients while columns are genes. However, without sufficient context, each value -- perhaps indicating the expression of a gene in a patient's tumor -- can be cryptic. Here, knowledge graphs (KGs) become invaluable for tabular data, as highlighted in PLATO~\cite{ruiz2022tabular}. These KGs are constructed from auxiliary domain-specific information. Using an medical example, a KG might be sourced from scientific literature, databases of genetic interactions, or annotations about gene functions. Within the KG, each gene (a feature) is represented as a node. This node might be connected to other genes, indicating interactions such as ``activates'' or ``inhibits.'' Additionally, it could link to nodes representing broader biological processes or functions, elucidating the gene's role. By integrating this structured knowledge from KGs, tabular data transforms from a mere matrix of values to a rich and interconnected web of insights, providing machine learning models with the depth they need to decode intricate patterns inherent in specialized domains.



\begin{table}[!t]
\centering
\caption{Summary of representative GNN models for graph representation learning on tabular data.}
\label{tab:gnnmodels}
\begin{tabular}{c|m{1.65cm}|m{5cm}}
\hline
Type & Model & References \\ \hline
\multirow{11}{*}{Ho.} & GCN & GINN~\cite{ginn20}, GEDI~\cite{gedi23}, IAGNN~\cite{iagnn21}, GCN-Int~\cite{gcnint21}, IDGL~\cite{idgl20}, RGSL~\cite{rgsl22}, DGM~\cite{dgm22tpami}, EGG-GAE~\cite{egggae23}, SLAPS~\cite{slaps21}, SUBLIME~\cite{sublime22}, TabGSL~\cite{tabgsl23}, LDS~\cite{lds19}, Table2Graph~\cite{table2graph}, FIVES~\cite{fives21}, IGNNet~\cite{ignnet23}, GNNBFD~\cite{xiao2023graph}, HES-GSL~\cite{hesgsl23}, LatentG~\cite{koloski2023latent} \\ \cline{2-3} 
 & GAT & GATE~\cite{gate20}, WPN~\cite{wpn21}, FinGAT~\cite{fingat21}, FT-GAT~\cite{ftgat23}, GNNDB~\cite{gnndb20}, GatFM~\cite{he2023novel} \\ \cline{2-3} 
 & GraphSAGE & LSTM-GNN~\cite{lstmgnn21}, GRAPE~\cite{you2020grape}, GNNDP~\cite{dpgnn21}, IGRM~\cite{igrm23aaai} \\ \cline{2-3} 
 & GIN~\cite{gin19} & DRSA-Net~\cite{drsanet23} \\ \cline{2-3} 
 & GAE~\cite{vgae16} & MST-GRA~\cite{gramst22}, GAEOD~\cite{du2022graph} \\ \cline{2-3} 
 & DGN~\cite{dgn18} & CCNS~\cite{ccns23neurips} \\ \cline{2-3} 
 & ProGNN~\cite{prognn20} & SSGNet~\cite{ssgnet22} \\ \cline{2-3} 
 & GGNN~\cite{ggnn16} & Fi-GNN~\cite{fignn19}, Causal-GNN~\cite{causalgnn23} \\ \cline{2-3}
 & Specialized & LUNAR~\cite{lunar22}, PET~\cite{pet22}, TabGNN~\cite{tabgnn21}, FATE~\cite{fate22}, SGANM~\cite{sganm22}, MGNN~\cite{mgnn22}, MedGraph~\cite{medgraph20}, T2G-Former~\cite{t2gformer23}, GNN4MV~\cite{gnn4mv22}, ALLG~\cite{allg22} \\ \hline
\multirow{6}{*}{He.} & GAT & GME~\cite{gme21}, GraphFC~\cite{graphfc23} \\ \cline{2-3}
 & GCN & LUCE~\cite{lifeprice23tkde} \\ \cline{2-3}
 & GraphSAGE & AO-GNN~\cite{aognn22} \\ \cline{2-3}
 & HAN~\cite{hgat19} & HSGNN~\cite{hsgnn20} \\ \cline{2-3}
 & HGT~\cite{hgt20} & xFraud~\cite{xfraud21}, RelBench~\cite{relbench} \\ \cline{2-3}
 & Specialized & GCT~\cite{gct20}, MAGNN~\cite{magnn22pr}, ST-GNN~\cite{yang2021financial}, C-FATH~\cite{cfath21}, GRC~\cite{xu2021towards}, H2FD~\cite{h2fd22}, PC-GNN~\cite{pcgnn21}, AMG-DP~\cite{amg20}, CARE-GNN~\cite{caregnn20}, RioGNN~\cite{riognn22}, RelBench~\cite{relbench} \\ \hline
 Hy. & Specialized & HCL~\cite{hcl22}, HyTrel~\cite{hytrel23}, PET~\cite{pet22} \\ \hline
\end{tabular}
\end{table}


\subsection{Representation Learning}
\label{subsec-replearn}
With the constructed graph for the given tabular dataset, the next stage is to accordingly produce the representation of each data instance through graph neural networks. Different GNN architectures are designed to capture various types of information and patterns from graphs. The choice of GNN often hinges on the specific characteristics and requirements of the derived tabular graph. We summarize which of the GNN methods is utilized by which work in Table~\ref{tab:gnnmodels}, which can be mainly divided into GNNs for the constructed homogeneous (Ho.) and heterogeneous (He.) graphs. In addition to directly employ or extend exsiting GNNs, some works develop GNNs specialized for tabular data with various key designs, summarized in Table~\ref{tab:specializedgnn}.

No matter homogeneous or heterogeneous graphs are constructed to depict the tabular data, the typical GNN architectures, i.e., GCN~\cite{gcn17}, GAT~\cite{gat18}, and GraphSAGE~\cite{graphsage}, are widely employed to learn instance representations. The reason is three-fold. The first is \textit{Expressiveness and Adaptability}. They can effectively capture and process both local and global information in a graph. This capability means they can be applied to a broad range of tabular datasets, making them adaptable to various tasks and scenarios~\cite{dlgsuv22,gnn21tnnls}. The second is \textit{Proven Performance}. Given their foundational nature in the realm of GNNs, they have been extensively benchmarked, refined, and validated across various tasks. Their consistent performance in numerous studies provides a level of assurance in their efficacy~\cite{gnn21tnnls,gnnsuv_aiopen,gnnsuv_trml}.
The third is \textit{Ease of Integration}. They can be easily integrated with other neural network components, allowing for the design of more complex models tailored to specific tasks. Furthermore, GAT~\cite{gat18} enables the model to weigh the importance of neighbors differently. For graphs converted from tabular data where certain relationships or nodes might be more relevant than others, GAT provides a more nuanced aggregation method. GraphSAGE~\cite{graphsage} handles vast graphs by sampling neighbors, offering scalability without compromising much on representation quality. If the tabular dataset is vast, resulting in a big graph, GraphSAGE can efficiently handle and learn representations by considering a sampled subset of neighbors.

\subsubsection{GNNs for Homogeneous Graphs}
ProGNN~\cite{prognn20} is constructed to handle noisy graphs and can refine the graph structure iteratively as it learns. For tabular datasets that might have been imperfectly converted into graphs, ProGNN can iteratively enhance both the graph structure and node representations~\cite{ssgnet22}. GGNN~\cite{ggnn16} utilizes gates to control the flow of information across nodes. If sequential aspects exist in the tabular dataset~\cite{causalgnn23}, or if there is a need to regulate the information flow in the graph more carefully~\cite{fignn19}, GGNN could be preferred. Due to the powerful structural discrimination ability, GIN~\cite{gin19} excel in instance graphs whose intricate relational patterns emerge based on tabular features. GIN can discern subtle differences in node relationships, capturing the interactions and combinations prevalent in tabular datasets~\cite{drsanet23}. To further capture high-order correlation between tabular elements, deeper models like DGN~\cite{dgn18} can be used to overcome the oversmoothing issue. The unsupervised nature of Graph AutoEncoder (GAE)~\cite{vgae16} allows it to distill the essence of a dataset by attempting to reconstruct its graph structure, making it suitable for capturing dense and interconnected relationships. In tabular datasets, the features are highly interconnected or exhibit patterns where capturing the joint distribution of multiple columns via GAE is crucial and effective~\cite{gramst22,du2022graph}.

\newcolumntype{C}[1]{>{\centering\arraybackslash}m{#1}}

\begin{table}[!t]
\centering
\caption{Summary of key designs on specialized GNNs.}
\label{tab:specializedgnn}
\resizebox{\columnwidth}{!}{
\begin{tabular}{C{2.9cm}|C{2.3cm}|C{2.3cm}}
\hline
Key Design & Homogeneous & Heterogeneous \\ \hline
Label Adjustment & PET~\cite{pet22}, SGANM~\cite{sganm22} &  \\ \hline
Feature-relation Modeling & TabGNN~\cite{tabgnn21}, ALLG~\cite{allg22} & GCT~\cite{gct20}, AMG~\cite{amg20}, CARE-GNN~\cite{caregnn20}, RioGNN~\cite{riognn22} \\ \hline
Feature Selection & T2G-Former~\cite{t2gformer23} & GRC~\cite{xu2021towards} \\ \hline
Neighbor Sampling &  & C-FATH~\cite{cfath21}, PCGNN~\cite{pcgnn21}, CARE-GNN~\cite{caregnn20}, RioGNN~\cite{riognn22} \\ \hline
Multi-modality & MGNN~\cite{mgnn22} & MAGNN~\cite{magnn22pr} \\ \hline
Missing Values & GNN4MV~\cite{gnn4mv22} &  \\ \hline
Temporal Dependency & MedGraph~\cite{medgraph20} & ST-GNN~\cite{yang2021financial} \\ \hline
Homo-Heterophilic & HES-GSL~\cite{hesgsl23} & H2FD~\cite{h2fd22} \\ \hline
Permutation Invariance & FATE~\cite{fate22} &  \\ \hline
Distance Preservation & LUNAR~\cite{lunar22} &  \\ \hline
\end{tabular}
}
\end{table}

\subsubsection{GNNs for Heterogeneous Graphs}
Heterogeneous Graph Attention Network (HAN)~\cite{hgat19}, which is inherently designed for heterogeneous structures, aligns perfectly with the multi-faceted relationships and diverse feature types present in tabular data transformed into heterogeneous graphs. Its attention mechanisms, both at node and semantic (meta-path) levels, ensure that instance representations capture the rich interactions and relationships inherent in such data~\cite{hsgnn20}.
Heterogeneous Graph Transformer (HGT)~\cite{hgt20} excels in processing heterogeneous graphs from tabular data due to its dynamic meta-path aggregation that captures intricate feature interactions. Its type-aware self-attention respects diverse feature distinctions, and its hierarchical attention ensures both granular and broad relationship comprehension, making it adept for tabular data's complexity~\cite{xfraud21}.

\subsubsection{Specialized GNNs}
Specialized GNNs for graphs derived from tabular data are crucial to tackle real data challenges, enhance performance, and meet various application purposes. Off-the-shelf  GNNs may miss tabular intricacies, while tailored models effectively capture nuanced relationships and handle data/application issues, ensuring optimal instance representations.

\textbf{Label Adjustment.} 
Both PET~\cite{pet22} and SGANM~\cite{sganm22} underscore the pivotal role of label-guided adjustments in refining tabular data graph representations. PET constructs a hypergraph where distinct feature values become nodes, facilitating message propagation for auxiliary label information transfer and feature space adjustments. SGANM employs a unique 2-D edge embedding, driven by label information, resulting in a self-adaptive graph structure. This adaptability, coupled with a multi-head masked attention mechanism, accentuates GNN's label-enhanced feature extraction capabilities. 

\textbf{Feature-relation Modeling.}
By taking each feature as a kind of relation to construct the instance graph, TabGNN~\cite{tabgnn21} can model finer-grained instance correlation with attention learning applied to each relation's graph to find which features contribute most to the prediction task.
To encode the dynamic instance correlation that evolves as the instance representations change in different neural network layers, ALLG~\cite{allg22} learns a series of relation propagated adjacent matrices so that more precise graph structures can be captured.
GCT~\cite{gct20} delivers a modification to the Transformer to guide the self-attention to learn the hidden conditional dependency structure between features using prior knowledge in the form of attention masks.
AMG-DP~\cite{amg20} presents a relation-specific receptive layer to leverage the local structure in the heterogeneous tabular graph, and to better incorporate rich semantics derived from multiplex feature relations.
CARE-GNN~\cite{caregnn20} and RioGNN~\cite{riognn22} formulate relation-aware neighbor aggregators to combine neighborhood information from different relations and produce instance embeddings.

\textbf{Feature Selection.}
By identifying and retaining only the most informative features, feature selection can reduce the graph's complexity, enhancing both computational efficiency and representation quality. This focused aspect mitigates noise, minimizes redundancy, and ensures that the graph structure effectively captures the underlying relationships in the data.
T2G-Former~\cite{t2gformer23} is a bespoke Transformer model for tabular learning with a Graph Estimator module for jointly selecting salient features and promoting heterogeneous feature interaction based on estimated relation graphs.
GRC~\cite{xu2021towards} characterizes the multiple types of relationships via self-attention mechanism and employs conditional random field to constrain nodes with the same type to select the focused features and produce similar representation.

\textbf{Neighbor Sampling.}
Given the diverse types of nodes and relations in a heterogeneous graph, neighbor sampling selectively aggregates information from a subset of relevant neighbors. This ensures efficient computation and prioritizes essential contextual information, enabling the model to capture the intricate relationships and heterogeneity intrinsic to the data, thereby enhancing the quality of learned instance representations.
By leveraging community tags, C-FATH~\cite{cfath21} addresses structure-inconsistency, ensuring nodes interact primarily with similar-behavior neighbors. Additionally, using entity similarities, C-FATH also tackles content-inconsistency, ensuring neighbors share features resembling the central node's.
In PCGNN~\cite{pcgnn21}, a label-balanced sampler is devised to pick nodes and edges for sub-graph training, and a neighborhood sampler is designed to choose neighbors for over-sampling the neighborhood of the minority class and under-sampling the neighborhood of the majority class. 
CARE-GNN~\cite{caregnn20} and RioGNN~\cite{riognn22} introduce similarity-aware neighbor selection mechanisms to identify neighbors closely related to a central node within a given relation in a heterogeneous graph. Utilizing reinforcement learning, they dynamically determine the optimal threshold for neighbor selection in tandem with the GNN training process.

\textbf{Multi-modality.}
GNNs are increasingly employed to process multimodal tabular data, harnessing the potential to capture intricate relationships and patterns across different data types. Through constructing appropriate graph representations, GNNs facilitate learning from disparate data modalities in an integrated manner.
MAGNN~\cite{magnn22pr} derives instance representations from a heterogeneous graph built on multimodal tabular inputs. Emphasizing model interpretability, MAGNN employs a dual-phase attention mechanism that jointly optimizes and elucidates the significance of both inner-modality and inter-modality connections.
MGNN~\cite{mgnn22} harnesses a unified GNN framework to extract features from multimodal medical datasets. By creating multiple bipartite graphs that map relationships between patients and various data sources, MGNN uses a GNN to ascertain each patient's embedding, along with a subsequent fusion layer to integrate these multimodal embeddings.

\textbf{Missing Values.}
GNN4MV~\cite{gnn4mv22} employs graph neural networks to classify instances in tabular data that have missing values, eliminating the need for imputation. By leveraging supervised information, it guides feature space construction, mitigating the influence of absent values when determining neighborhood relationships. The neighborhood graph convolutional mechanism within GNN4MV uses the graph's inherent structure to directly classify these incomplete instances.

\textbf{Temporal Dependency.}
In domains like electronic medical records and financial user-item interactions, tabular data often contains temporal nuances. GNNs adeptly capture these temporal dependencies across instances and features. MedGraph~\cite{medgraph20} presents a method to seamlessly grasp both the structural co-location information between patient visits and codes, and the chronological sequencing of visits. Leveraging a visit-code bipartite graph, MedGraph integrates temporal point processes, offering an end-to-end solution for capturing medical history.
ST-GNN~\cite{yang2021financial} processes temporal tabular graphs, capturing structural embeddings for nodes across snapshots. Using a temporal-aware aggregator, it derives time-sensitive embeddings. Combining these via attention, ST-GNN delivers a node embedding integrated with the end-to-end learning objective.

\textbf{Homo-Heterophilic.}
In tabular data prediction models, instances often exhibit complex correlation with others, leading to the establishment of both \textit{homophilic} (similar nodes) and \textit{heterophilic} (dissimilar nodes) connections. Most GNNs predominantly emphasize homophily in the data graph~\cite{hesgsl23}, using low-pass filters to maintain node feature commonalities among neighbors. This approach can inadvertently overlook the nuances in heterophilic connections. H$^2$-FD~\cite{h2fd22} distinguishes between homophilic and heterophilic connections with guidance from labeled nodes. Its information aggregation strategy ensures homophilic connections propagate analogous information, while heterophilic ones highlight distinct information.

\textbf{Permutation Invariance.}
Permutation invariance with respect to the order of input features is a critical property for tabular data modeling. Essentially, it ensures that the model's output remains consistent regardless of the sequence in which input features are presented. This property is crucial because, in many real-world datasets, the order of features is arbitrary and should not influence the model's prediction. Moreover, permutation invariance inherently provides flexibility to handle variable-length input feature vectors~\cite{deepsets17}. As new features emerge, either from previously unseen values of existing features or entirely new raw features, the model remains robust and adaptable. FATE~\cite{fate22} presents a permutation-invariant aggregation in the tabular GNN model. The idea is to replace the embedding layer with an embedding lookup and a sum aggregation over indexed feature embeddings. 

\textbf{Distance Preservation.}
Distance between instances is pivotal in tabular data learning tasks, especially in anomaly detection and clustering where relative proximities among data points provide significant insights. LUNAR~\cite{lunar22} incorporates these distances by representing them as edge features within a GNN framework. In the GNN's message aggregation phase, these edge features, which encapsulate the distance information, are encoded. As a result, the derived instance representations inherently embody the distance-based relationships, offering a richer, more nuanced understanding of the tabular data's structure. The distance-preserving tabular GNNs improve generalization due to their ability to recognize the relative positioning of instances, bolstering robustness against noisy data.


\begin{table}[!t]
\centering
\caption{Summary of additional learning tasks (besides the main task of tabular data prediction).}
\label{tab:learntasks}
\resizebox{\linewidth}{!}{%
\begin{tabular}{c|l}
\hline
Learning Tasks & Representative studies \\ \hline
\multirow{2}{*}{Feature Reconstruction} & GINN~\cite{ginn20}, GEDI~\cite{gedi23}, EGG-GAE~\cite{egggae23}, IGRM~\cite{igrm23aaai}, \\
                                        & MST-GRA~\cite{gramst22}, GAEOD~\cite{du2022graph}, ALLG~\cite{allg22}, GRAPE~\cite{you2020grape} \\ \hline
Denoising Autoencoder & SLAPS~\cite{slaps21}, HES-GSL~\cite{hesgsl23} \\ \hline
Contrastive Learning & SUBLIME~\cite{sublime22}, TabGSL~\cite{tabgsl23}, SSGNet~\cite{ssgnet22} \\ \hline
Graph Regularization & IDGL~\cite{idgl20}, MST-GRA~\cite{gramst22}, GraphFC~\cite{graphfc23}, ALLG~\cite{allg22} \\ \hline
Sparsity Regularization & Table2Graph~\cite{table2graph} \\ \hline
Multi-level Regularization & ALLG~\cite{allg22} \\ \hline
\multirow{2}{*}{Property Preservation} & GINN~\cite{ginn20}, LUCE~\cite{lifeprice23tkde}, GCT~\cite{gct20}, \\ & MedGraph~\cite{medgraph20}, TabularNet~\cite{tabularnet21} \\ \hline
Explanation Preservation & xFraud~\cite{xfraud21} \\ \hline
\end{tabular}%
}
\end{table}

\begin{table}[!t]
\centering
\caption{Summary of learning strategies.}
\label{tab:learnstg}
\resizebox{\linewidth}{!}{%
\begin{tabular}{c|l}
\hline
Strategy & Representative studies \\ \hline
\multirow{3}{*}{Two-stage} & SUBLIME~\cite{sublime22}, GNNBFD~\cite{xiao2023graph}, GRAPE~\cite{you2020grape}, \\ 
                           & IGRM~\cite{igrm23aaai}, GINN~\cite{ginn20}, MST-GRA~\cite{gramst22}, \\
                           & GAEOD~\cite{du2022graph}, MedGraph~\cite{medgraph20} \\ \hline
Adversarial & GINN~\cite{ginn20} \\ \hline
Alternative & GEDI~\cite{gedi23} \\ \hline
Bi-level & LDS~\cite{lds19}, FIVES~\cite{fives21}, FATE~\cite{fate22} \\ \hline
Pretrain-finetune & ALLG~\cite{allg22}, GraphFC~\cite{graphfc23} \\ \hline
\multirow{3}{*}{End-to-end} & Most of the rest methods \\
 & (e.g., TabGSL~\cite{tabgsl23}, T2G-Former~\cite{t2gformer23}, LUNAR~\cite{lunar22} \\
 & TabGNN~\cite{tabgnn21}, PET~\cite{pet22}, DGM~\cite{dgm22tpami}, Fi-GNN~\cite{fignn19}) \\ \hline
\end{tabular}%
}
\end{table}

\subsection{Training Plans}
\label{subsec-trainplan}
Crafting effective training plans, encompassing both learning tasks and strategies, is essential in GNN-based tabular data learing. The learning tasks, such as feature reconstruction or graph regularization, tailor GNNs to intricately capture the complex relationships within tabular data. Concurrently, the selection of learning strategies, like bi-level optimization or pretrain-finetune methods, critically influences the model's generalization, robustness, and adaptability. This deliberate orchestration of training plans is fundamental; it ensures that GNNs not only generate precise and interpretable embeddings but also enhance predictive accuracy and reliability. Such well-structured training plans are crucial for leveraging GNNs to their fullest potential in deriving meaningful insights from tabular datasets across diverse applications. We have created Table~\ref{tab:learntasks} and Table~\ref{tab:learnstg} to summarize various aspects of learning tasks and strategies, respectively.

\subsubsection{Learning Tasks}
\textbf{Representation Enahacement}
through learning to reconstruct in GNN-based tabular data learning involves training the model to not only learn effective representations but also to accurately reconstruct the original features. This process, mainly involving either feature reconstruction or contrastive learning, enhances model robustness by encouraging the GNN to focus on and preserve key features during representation learning. 
\begin{itemize}[leftmargin=*]
\item \textit{Feature Reconstruction.}
Reconstructing features from embeddings obtained through a GNN encoder essentially involves decoding the rich, condensed information captured in the embeddings back into the original feature space. It serves as a robustness check, ensuring that the GNN encoder captures all relevant information and patterns present in the data. This reconstruction acts as a form of regularization, preventing the model from overfitting by learning to preserve essential data characteristics. There are three approaches: leveraging complete feature sets for high-fidelity representation, handling missing values to enhance model robustness, and processing noisy data to improve resilience against data quality issues. (a) \textbf{Complete Features}: GNNs trained on complete feature vectors produce detailed representations, ensuring comprehensive data integrity~\cite{du2022graph,allg22}. This approach is ideal for applications requiring exact feature preservation. (b) \textbf{Missing Values}: GNNs learn from incomplete data, making them adept at handling real-world datasets with gaps~\cite{you2020grape,gedi23,egggae23,igrm23aaai}. The reconstructed features can effectively impute missing values, enhancing the model's robustness in imperfect data scenarios. (c) \textbf{Noisy Data}: Training on noisy features through denoising autoencoder allows GNNs to become resilient to data corruption, focusing on underlying data patterns~\cite{ginn20,slaps21,gramst22,hesgsl23}. This is beneficial for maintaining consistent performance even with compromised data quality.

\item \textit{Contrastive Learning.} 
Contrastive learning with GNN encoder embeddings boosts tabular data learning by refining the model's ability to differentiate between similar and dissimilar instances. This approach enhances representation quality, and bolsters the model's generalizability, ensuring it focuses on fundamental data characteristics and avoids overfitting to specific instances. SUBLIME~\cite{sublime22} optimizes graph topology using contrastive learning to create robust structures in unsupervised settings. TabGSL~\cite{tabgsl23} applies contrastive learning to enhance instance correlation and feature interaction, improving classification tasks. SSGNet~\cite{ssgnet22} employs a siamese GNN architecture with a contrastive objective for instance similarity learning in medical tabular data, addressing sparsity and missing values. Each method utilizes contrastive learning to refine graph structures and node embeddings
\end{itemize}

\textbf{Regularization.} 
Regularization on embeddings from GNN encoders is a critical aspect of GNN-based tabular data learning, ensuring the stability and effectiveness of the learned graph structures and the learned instace representations. There are three primary types of regularization. The first is \textit{Graph Regularization}: maintaining balance in connectivity to avoid overly smooth or sparse graphs. Representative regularizers include reducing adjacent nodes' embeddings~\cite{idgl20,graphfc23,allg22}, and minimizing Minimum Spanning Tree (MST) distance~\cite{gramst22} that preserves local neighborhood structure between instaces.  The second is \textit{Sparsity Regularization}, as employed by Table2Graph~\cite{table2graph}, controls the adjacency matrix's sparsity to highlight key feature interactions while avoiding overfitting. The third is \textit{Multi-level Regularization}, like in ALLG~\cite{allg22}, uses multiple levels of adjacency matrices, with each layer informed and regularized by its predecessor, allowing for smooth propagation of relationships among samples. 

\textbf{Information Preservation.}
The learning task of information preservation in embeddings focuses on retaining essential characteristics and relationships inherent in the tabular dataset, ensuring the model's output accurately reflects these core aspects. There are two main types of information being preserved: \textit{Property Preservation} and \textit{Explanation Preservation}. Property Preservation, as seen in models like GINN~\cite{ginn20}, GCT~\cite{gct20}, and LUCE~\cite{lifeprice23tkde}, aims to maintain global dataset attributes, conditional probabilities between features, and geographical distances among instances, respectively, in the learned representations. In addition, 
MedGraph aims at preserving medical dependency between disease and diagnosis~\cite{medgraph20}, and TabularNet~\cite{tabularnet21} targets at preserving spatial information encoding on some features~\cite{tabularnet21}.
On the other hand, Explanation Preservation focuses on ensuring that explanations highlighted by domain experts are adequately represented in the GNN's subgraph explanations~\cite{xfraud21} produced by GNNExplainer~\cite{gnnexplainer19}. These preservation tasks are crucial for ensuring the GNN's embeddings are not just accurate but also meaningful and interpretable in the context of the original tabular data.


\subsubsection{Training Strategies}
The strategies range from structured approaches, prioritizing sequential learning and optimization, to unified methods, emphasizing comprehensive model development. Below we review six widely adopted strategies.

\begin{itemize}[leftmargin=*]
\item \textbf{Two-stage} methods involve a sequential approach where the initial task is to learn either a graph structure or instance representations from the tabular data. Subsequently, these learned structures or embeddings are utilized in training downstream prediction models. For instance, SUBLIME~\cite{sublime22} first learns an instance graph and representations from tabular data, then employs these in training a GNN model for predictions. Similarly, models like GRAPE~\cite{you2020grape}, GINN~\cite{ginn20}, and IGRM~\cite{igrm23aaai} focus initially on imputing missing values using tabular GNNs, followed by leveraging these imputed features for downstream predictions. Other approaches, such as GNNBFD~\cite{xiao2023graph}, MST-GRA~\cite{gramst22}, GAEOD~\cite{du2022graph}, and MedGraph~\cite{medgraph20}, generate instance representations through GNN-based encoders built on various graphs derived from tabular data, which are then used in training models for final predictions. This separation allows for targeted optimization at each stage, potentially leading to enhanced model performance. However, drawbacks include increased computational demands and a potential disconnect between stages, where improvements in initial representation learning may not directly translate to better prediction outcomes. Additionally, the need for effective integration between stages can add to the complexity of the process.

\item \textbf{Adversarial} methods in GNN-based tabular data predictions, exemplified by approaches like GINN~\cite{ginn20}, incorporate adversarial training strategies to enhance the model's ability to discern between imputed and real data. This technique involves using an adversarial loss during feature reconstruction, compelling the reconstructed vectors to closely align with the natural distribution of the original data patterns. While this method improves the realism and accuracy of predictions, it also adds complexity and computational demands to the training process, potentially impacting model convergence.

\item \textbf{Alternative} methods like that used in GEDI~\cite{gedi23} involve adaptively optimizing the weights of feature reconstruction tasks based on the performance of the target task. It treats feature reconstruction weights as meta-parameters, allowing for a more dynamic and responsive training process. This approach enhances the reconstruction of crucial features under the guidance of downstream tasks. The adaptive weighting mechanism guards against negative transfer from multi-task learning, ensuring auxiliary tasks do not hinder the primary objective. However, this method adds complexity in managing meta-parameters, requiring careful balancing for optimal task performance.

\item \textbf{Bi-levevl} optimization involves structuring training as two inner-outer linked optimization problems. Models like LDS~\cite{lds19} and FIVES~\cite{fives21} utilize this approach to learn a discrete and sparse dependency structure among data points and features, respectively, while concurrently training a GCN's parameters. This method treats GCN parameter training as the 'outer' optimization problem, guided by a generative probabilistic model for graphs. The dual focus on optimizing both GCN parameters and the graph generator aims to minimize classification errors on the dataset. Similarly, FATE~\cite{fate22} employs strategies like proxy training data and asynchronous updates to enhance model extrapolation for new features. This involves using partial features for updates and decoupling the training of the prediction and GNN networks, updating them at different intervals.

\item \textbf{Pretrain-Finetune} methods involves a two-step approach, exemplified by models like GraphFC~\cite{graphfc23} and ALLG~\cite{allg22}. In the pre-training stage, the model undergoes self-supervised learning on local neighborhood preservation or feature reconstruction, utilizing both labeled and unlabeled data to understand feature distributions and instance correlations. Subsequently, in the finetuning stage, the model weights are refined using labeled data, focusing on target label predictions. This approach benefits from establishing a robust initial understanding of the data, leading to more accurate predictions after finetuning. However, a potential drawback is the possible mismatch between pre-training and finetuning stages, especially if the data distributions vary. 

\item \textbf{End-to-end} training is a prevalent method in GNN-based tabular data predictions, streamlining the process from representation learning to final prediction in a unified model. Models can be exampled (but not limited) by those listed in Table~\ref{tab:learnstg}. This approach is favored for its efficiency and ability to optimize the entire GNN for specific prediction tasks, often leading to improved accuracy by capturing intricate data relationships. However, it can be prone to overfitting, particularly with limited data, and may require substantial computational resources. Despite these challenges, end-to-end training's comprehensive and cohesive nature makes it a popular choice for effectively harnessing GNNs in tabular data learning.
\end{itemize}


\section{Applications}
\label{sec-app}
The versatility of GNN4TDL has been useful across various domains. GNNs excel in complex tasks like anomaly detection (Sec.~\ref{subsec-fraud}), click-through rate prediction (Sec.~\ref{subsec-ctr}), and medical prediction (Sec.~\ref{subsec-med}), showcasing their ability to untangle intricate data relationships. Their profound impact extends to the nuanced challenges of missing data imputation (Sec.~\ref{subsec-impute}) and relational database modeling (Sec.~\ref{subsec-db}), redefining traditional approaches in these critical areas.

\subsection{Anomaly Detection}
\label{subsec-fraud}
On the task of anomaly detection for tabular data, GNNs have emerged as a powerful tool, addressing the limitations of traditional statistical, supervised learning, and distance-based methods. GNNs excel in capturing complex dependencies and handling relational data, a critical advantage over traditional methods which often make strong assumptions about data distribution and fail to capture intricate feature interactions. For example, LUNAR \cite{lunar22} integrates local outlier methods into GNNs, enhancing adaptability and expressivity in anomaly detection. Similarly, graph autoencoders \cite{du2022graph, gramst22} have proven effective in altering data distribution patterns to identify embedded outliers.
In fraud detection, closely related to anomaly detection, GNNs adeptly uncover fraudulent activities by exploiting data connections. Models such as CARE-GNN \cite{caregnn20}, PC-GNN \cite{pcgnn21}, AO-GNN \cite{aognn22}, H2-FDetector \cite{h2fd22}, IHGAT \cite{ihgat21kdd}, xFraud \cite{xfraud21}, C-FATH \cite{cfath21}, HMF-GNN \cite{zhang2022hierarchical}, GraphBEAN~\cite{graphbean23}, and GraphFC \cite{graphfc23} demonstrate GNNs' capabilities in detecting fraud through both homogeneous and heterogeneous graph-based approaches.
In fault detection, GNNs like IAGNN \cite{iagnn21}, SGANM \cite{sganm22}, and GNNBFD \cite{xiao2023graph} showcase their effectiveness in identifying faults or failures in systems, such as mechanical equipment and industrial processes. These GNNs leverage the interaction among sensor signals and sample relationships to enhance fault detection, exemplifying their superior capability in identifying deviations in complex systems.

\subsection{CTR Prediction}
\label{subsec-ctr}
In the application of Click-Through Rate (CTR) prediction within the GNN4TDL framework, GNNs address critical challenges that traditional models face, such as sparse data, complex feature interactions, and user-item heterogeneity. Traditional approaches like logistic regression and deep learning models (DeepFM \cite{deepfm17}, Wide\&Deep \cite{wnd16}, GCN-int \cite{gcnint21}) struggle with these issues, which GNNs effectively overcome. GNNs leverage structural information in user-item interaction graphs to mitigate sparse data and cold-start problems, as demonstrated by GME \cite{gme21}, which constructs graphs connecting old and new ads. For complex feature interactions, GNNs like Fi-GNN \cite{fignn19} and Cross-GCN \cite{crossgcn21} model interactions between features as node attributes in the graph, capturing high-order interactions more efficiently. In addressing user and item heterogeneity, GNNs employ varied node and edge types to represent different user and item categories, as seen in DisenCTR \cite{disenctr22} and DE-GNN \cite{dgenn21}, which utilize time-evolving graphs and dual graph structures, respectively. Additionally, in the context of sequential recommendation, RetaGNN \cite{retagnn21} leverages a user-item-attribute tripartite graph, enhancing the capability of recommendation systems in diverse settings. This multifaceted approach by GNNs in CTR prediction showcases their superiority in capturing the nuanced dynamics of user-item interactions, significantly improving prediction accuracy in online services.

\subsection{Medical Prediction}
\label{subsec-med}
For medical prediction, GNNs have significantly advanced the analysis and prediction capabilities within healthcare, particularly in handling Electronic Medical Records (EMRs). EMRs, rich with patient visit details, diagnostic codes, and treatment information, present challenges such as structural and temporal complexity, data sparsity, and heterogeneity. GNNs address these challenges effectively, as demonstrated in various applications including disease prediction \cite{dpgnn21,wpn21,riognn22}, patient representation learning \cite{riognn22,hsgnn20,ssgnet22,hcl22,medgraph20,gct20}, medication recommendation \cite{gate20}, and survival prediction \cite{lstmgnn21,riognn22,mgnn22}. They enhance the understanding of the intrinsic structure and temporal dynamics in EMRs by treating medical codes as nodes, thus capturing the complex interrelations and temporal effects. In tackling data sparsity, particularly for rare diseases, GNNs excel by leveraging external knowledge and interconnected nodes, allowing for more accurate predictions and generalization. Additionally, the inherent heterogeneity of EMRs, with diverse patient information and medical histories, is adeptly managed by GNNs, which can interpret and utilize this complex, graph-like data structure. The application of GNNs in medical prediction thus marks a significant leap in exploiting the nuanced and intricate nature of healthcare data, leading to more informed and precise medical outcomes.

\subsection{Missing Data Imputation}
\label{subsec-impute}
GNNs have emerged as a transformative solution in the field of missing data imputation for tabular data, addressing key challenges that traditional imputation methods face. Traditional techniques, ranging from simple mean or median imputation to more complex machine learning models like autoencoders \cite{ginn20,egggae23}, often struggle with issues such as non-random missingness, multivariate missingness, and the disconnect between imputation and target prediction. GNNs, with their ability to model intricate data dependencies, provide a more nuanced approach to these challenges. For instance, GNN4MV \cite{gnn4mv22} tackles non-random missingness by using a graph-based classification approach, leveraging neighbor information to guide the construction of a reliable graph structure. In addressing multivariate missingness, GNNs like GEDI \cite{gedi23} and GRIN \cite{grin22} effectively capture inter-feature relationships, with GEDI projecting features into a common space using an attention mechanism, and GRIN learning spatio-temporal representations in a multivariate context. Furthermore, GRAPE \cite{you2020grape} and IGRM \cite{igrm23aaai} innovatively integrate imputation with target prediction, treating imputation as an edge weight prediction task and employing a friend network to enhance the differentiation of sample relationships. This comprehensive approach by GNNs in missing data imputation significantly enhances the accuracy and applicability of imputation methods across various domains, from healthcare and finance to environmental science and social sciences, ensuring more robust and reliable predictive modeling in datasets with incomplete information.

\subsection{Financial Technology}
\label{subsec-fintech}
GNNs are revolutionizing the handling and analysis of tabular financial data, such as transaction records, credit histories, and stock prices. GNNs can model complex interdependencies that traditional tabular data processing techniques overlook. For instance, in asset price prediction~\cite{lifeprice23tkde,magnn22pr,rgsl22,tgc19tois}, GNNs provide a nuanced understanding of the stock market by recognizing the interconnected nature of stocks, sectors, and market indicators, offering more context-rich predictions than traditional methods. This approach is critical in scenarios where the movement of one stock, like a major tech company, can significantly influence related stocks in the sector.
In credit scoring \cite{amg20}, GNNs transcend the limitations of treating each client as an isolated entity by modeling the shared transactions and business relationships, thus uncovering hidden patterns in borrower networks that can inform credit risk assessments and management more accurately~\cite{yang2021financial,sgrn22ijcai}. Similarly, in investment decision-making \cite{fingat21,kim2019hats}, GNNs utilize their ability to interpret complex relationships between various financial entities, enhancing decision-making processes with deeper insights. Moreover, GNNs are effective in detecting financial fraud~\cite{xu2021towards,graphfc23}, where they address the challenge of coordinated activities among multiple accounts. By modeling the interactions among different accounts, GNNs are capable of identifying complex fraudulent activities that traditional methods might miss. This ability to capture and analyze the intricate web of financial transactions and relationships demonstrates the impact of GNNs in FinTech, paving the way for more sophisticated, accurate, and reliable financial analyses and decision-making processes.

\subsection{Modeling Relational Databases}
\label{subsec-db}
The modeling of relational databases can be enhanced through GNNs by capitalizing on the inherent structure and relationships within database schemas. This approach marks a significant departure from traditional database handling, which often involves flattening data tables and extensive feature engineering. As exemplified by RDBToGraph \cite{gnndb20}, relational databases are interpreted as directed multigraphs, where nodes represent database rows and edges denote relationships defined by keys and references. This structure effectively preserves the relational context that is typically lost in conventional database processing methods.
Siamese Graph Convolutional Networks (S-GCNs) \cite{krivosheev2020siamese} integrate both structured and unstructured data sources, such as relational databases and free text. By constructing graphs where records are nodes and their relations are edges, and linking these with graphs derived from unstructured text, S-GCNs facilitate a comprehensive understanding of data relationships, enhancing tasks like record linkage.
Combining language models (LMs) with GNNs can beeter exploit the full relational structure of databases~\cite{vogel2022towards}. This approach involves using LMs to encode individual rows and GNNs to model relationships within and across tables.
Lastly, RelBench~\cite{relbench} introduces Relational Deep Learning, an end-to-end approach that views relational tables as heterogeneous graphs. This method allows GNNs to learn across multiple tables, automatically extracting representations that leverage all available data without manual feature engineering.

\begin{table*}[!t]
\centering
\caption{Comparison of three ways to use features in terms of their advantages and disadvantages.}
\label{tab-feattable}
\begin{tabular}{C{2cm}|C{7cm}|C{7cm}}
\hline
\textbf{} & \textbf{Pros} & \textbf{Cons} \\
\hline
Used as \textbf{feature nodes} &
\begin{itemize}[leftmargin=*]
    \item Explicitly model the relationship between instances and their features, making it easier for GNNs to capture feature-level interactions.
    \item Preserve the original tabular data structure, which can be advantageous for certain learning needs, such as handling feature heterogeneity (e.g., continuous, categorical, ordinal) and integrating metadata or external knowledge.\vspace{-1em}
\end{itemize} &
\begin{itemize}[leftmargin=*]
    \item May not efficiently capture complex relationships between instances, as the graph structure only models relationships through shared features.
    \item Heterogeneous information might require tailored message-passing mechanisms to handle different types of nodes and edges.\vspace{-1em}
\end{itemize} \\
\hline
Used to \textbf{create edges} &
\begin{itemize}[leftmargin=*]
    \item Provide a richer representation of relationships between instances based on features, enabling GNNs to effectively learn complex patterns.
    \item Can capture higher-order relationships between instances by considering feature interactions.\vspace{-1em}
\end{itemize} &
\begin{itemize}[leftmargin=*]
    \item These features cannot be used for neighborhood aggregation to produce instances’ representations.
    \item Need to choose manually determine the similarity metrics for different features to connect instances.\vspace{-1em}
\end{itemize} \\
\hline
Used as \textbf{initial vectors} &
\begin{itemize}[leftmargin=*]
    \item Integrate feature information directly into node attributes, making it accessible for GNN learning.
    \item Compatible with various GNN architectures, allowing for more flexible model design.\vspace{-1em}
\end{itemize} &
\begin{itemize}[leftmargin=*]
    \item May not capture feature-level relationships and dependencies between instances as effectively.
    \item Can result in limited interpretability, as harder to reason about feature importance or their influence on predictions due to the implicit nature of features in the node vectors.\vspace{-1em}
\end{itemize} \\
\hline
\end{tabular}
\end{table*}

\section{Open Problems \& Future Directions}
\label{sec-directions}

\textbf{Obtaining the Ability of Tree-based Models.} 
A recent study~\cite{grinsztajn2022why} had discussed the potential reasons that tree-based models still outperform deep learning on typical tabular data. Key findings highlight that neural networks struggle to create best-fit functions for non-smooth decision boundaries, while tree-based methods do much better with irregular patterns. In addition, irrelevant features can significantly degrade the performance of neural network models. But tree-based models have the ability to stay insulated from the effects of worse features. While developing the GNN models for tabular data learning, the performance can be boosted if we can incorporate the abilities to learn weird patterns and better select useful features like tree-based methods. Recent advances have effectively attempted to transfer such abilities from tree-based models to neural networks (e.g., NBDT~\cite{wan2021nbdt}, DeepGBM~\cite{deepgbm19}, and ANTs~\cite{ants19}). How to have such designs in GNNs for tabular data learning would be a useful direction.

\textbf{Incorporating Graph Transformers.} Standard transformers with graph-specific modifications can learn effective representations of node and edge tokens~\cite{kim2022pure}. We have seen that various elements in tabular data can be treated as nodes, and their correlation can be utilized as edges. An appropriate arrangement of tabular elements as the input of transformers will not only bring the advantages of GNNs discussed in Section~\ref{subsec-whygnntdl}, but also impose the strength of transformers. The representation quality of data instances can be further improved because transformers are capable of learning complex representations and interactions between features in the data. Besides, we will have a natural way to handle missing values since transformers can implicitly learn to handle missing values through the self-attention mechanism, which assigns low attention scores to the missing data points. Even one can change the models of graph representation learning, described in Section~\ref{subsec-replearn}, to graph-enhanced transformers like Structure-Aware Graph Transformer~\cite{sat22icml} and GPS Graph Transformer~\cite{rampasek2022recipe}, so as to enjoy the advantages of transformers in tabular data learning.

\textbf{Scaling GNNs to Large Tabular Data.}
Tabular data in real-world applications, such as click-through rate prediction~\cite{ltu21cikm} and fraud detection~\cite{date20kdd}, may contain millions of instances and thousands of features. Formulating and constructing various graphs with their representation learning for large-scale tabular data requires huge computational costs, which make existing approaches infeasible. Three different strategies can bring scalability into GNN4TDL. The first is to choose a compact graph formulation that requires a relatively small number of nodes and/or edges constructed for tabular data. Hypergraphs can be a computationally efficient formulation~\cite{hgnn23tpami}. The second is to apply sparse learning techniques to produce graph sparsification that samples a subgraph to reduce the amount of data aggregation and to achieve model sparsification that prunes the neural network to reduce the number of trainable weights~\cite{zhang2023joint}. The third is to adopt scalable GNN models, such as PPRGo~\cite{pprgo20kdd}, NDLS~\cite{zhang2021node}, GraphAutoScale~\cite{fey2021gnnautoscale}, and GraphFM~\cite{yu2022graphfm}. 

\textbf{Non-homogeneous Graph Structure Learning for Tabular Data.}
We have seen a variety of learning-based approaches to construct the graph structures that depict tabular data in Section~\ref{subsec-graphcons}. However, most of them focus on learning homogeneous graphs, in which nodes are all data instances. The learning of graph structures for bipartite graphs, heterogeneous graphs, and hypergraphs constructed from tabular data is fully unexplored. While edges in homogeneous graphs depict the correlation between instances or between features, the learning of non-homogeneous graph structures of tabular data involves connecting data instances with various features. The underlying meanings can be elaborated from multiple aspects. For example, a learned edge can be viewed as associating an instance with an additional feature, which is essentially the effect of data augmentation. Since some instances can contain missing feature values, connecting them to features is a kind of data imputation, which is also the self-supervised learning task on tabular data. Furthermore, 
since graph structure learning considers not only edge addition but also edge removal, eliminating the association between instances and features, which is equivalent to creating missing values, can be regarded as adversarial learning, which can improve the generalization ability and the model robustness for tabular data prediction.


\textbf{Better Strategies to Utilize Features.}
When representing tabular data as a graph, the common formulations are instance graphs and bipartite graphs, which require features to create edges and nodes, respectively. When applying graph neural networks, we need the initial vectors associated with nodes for information propagation. Features of an instance have multi-way usages -- used as feature nodes (in bipartite graphs), used to create edges between instance nodes (in instance graphs), and used as the initial vectors of instance nodes (in both bipartite and instance graphs). We summarize the advantages and disadvantages of such three usages of features in Table~\ref{tab-feattable}. It is still unexplored to investigate which kinds of tabular features better fit which types of usages for GNNs. For example, which features are better used in initial node vectors? Also, which features should be used to create edges? To unleash the power of GNNs for tabular data, one can also seek to optimize the feature usage. The task is to select subsets of features for different usages in either a disjoint or overlapping manner so that the prediction performance can be boosted. This direction is also related to feature selection on graph neural networks, in which not utilizing irrelevant features as initial node vectors can improve the performance of node classification~\cite{xie22fea2fea}. The selected features as initial vectors also cannot be redundant as highly correlated features would cause the performance degradation of deep GNNs~\cite{decorr22kdd}.

\textbf{Graph-based SSL for Tabular Data.}
Self-supervised learning (SSL) has been proven to be useful in deep learning-based tabular data prediction. Typical auxiliary SSL tasks applied to tabular data include contrastive learning~\cite{subtab21,scarf22}, feature reconstruction~\cite{vime20,tabnet21}, data imputation~\cite{you2020grape}, and column prediction~\cite{nam2023stunt}. While tabular data is represented by a graph, one can further leverage the structure knowledge depicting the correlation between instances (and/or features), together with original features, to design proper graph-based SSL tasks~\cite{gssl22}. Here are some potential SSL tasks based on tabular graphs. (a) \textit{Missing Feature Imputation}: Predict missing feature values by training the GNN to reconstruct the known features for each instance. This helps the model learn feature-level relationships and dependencies. (b) \textit{Graph Clustering}: Learn node embeddings that group similar instances together by optimizing a clustering objective (e.g., maximizing intra-cluster similarity and minimizing inter-cluster similarity). (c) \textit{Graph Completion}: Train the GNN to predict missing edges in the graph based on the existing edges and node attributes, learning to capture higher-order relationships between instances. (d) \textit{Neighborhood Prediction}: Train the GNN to predict the neighbors of a given node based on the node's attributes and graph structure, learning to recognize local patterns and relationships. (e) \textit{Denoising or De-corrupting Graphs}: Train the GNN to reconstruct the original graph from a noisy or corrupted version by optimizing a graph reconstruction loss, learning to capture robust and clean representations of instances. (f) \textit{Contrastive learning}: Create positive and negative pairs of instances (e.g., based on feature similarity or other criteria) and train the GNN to distinguish between them, learning informative and discriminative representations. The SSL tasks help GNNs learn effective and expressive representations from tabular data, which can then be fine-tuned for downstream tasks such as classification, regression, and recommendation.

\textbf{Dealing with Robustness Issues.}
Applying GNNs for tabular data prediction introduces robustness issues that arise from different factors. Below, we discuss these robustness issues that require further exploration in designing GNNs for tabular data. (a) \textit{Noise in Graph Structure}: Spurious edges or incomplete connections, which result from noisy features and missing values, can hinder the model's ability to learn and generalize effectively~\cite{rsgnn22wsdm}, as they can lead to the incorrect propagation and aggregation of information in the GNN. (b) \textit{Data Distribution Shifts}: GNNs may struggle with shifts in the data distribution, such as changes in feature distributions or relationships between instances~\cite{wu2022handling}. While GNNs can capture complex patterns in the training data, they might not generalize well to unseen data with different characteristics. (c) \textit{Overfitting and Oversmoothing}: GNNs can suffer from overfitting, especially when learning from small tabular datasets. Oversmoothing, a phenomenon where node representations become excessively similar after multiple layers of aggregation, can further exacerbate this issue, reducing the model's ability to discriminate between instances~\cite{Zhao2020PairNorm}. The way graphs are constructed can affect the degree of overfitting and oversmoothing. (d) \textit{Adversarial Attacks}: GNNs can be vulnerable to adversarial attacks, where small perturbations to the graph structure or node features are introduced to mislead the model~\cite{li2023revisiting}. Such attacks can exploit the model's sensitivity to the graph structure and feature noise, potentially causing significant performance degradation. A tabular GNN model needs to be robust to structural perturbations that come from maliciously crafted feature values on instances.


\section{Conclusions \& Discussion}
\label{sec-conclude}
This survey paper provides a comprehensive examination of Graph Neural Networks (GNNs) for tabular data learning, contributing significantly to the understanding and advancement of this critical research area. Its primary contribution lies in its extensive exploration of the underlying processes -- graph formulation, graph construction, graph representation learning, and training plans -- which serves as a detailed guide for both novice and experienced researchers. In addition, by showcasing a broad range of practical applications of tabular GNNs, this paper substantiates the versatility and adaptability of GNNs in real-world scenarios. Furthermore, our delineation of potential future research directions fills a crucial gap in the existing literature, providing a roadmap for those keen on advancing this technology. By shedding light on these aspects, the paper serves as an invaluable resource, fostering understanding, accelerating innovations, and shaping the future trajectory of research in GNNs for tabular data learning.

\ifCLASSOPTIONcompsoc
  \section*{Acknowledgments}
\else
  \section*{Acknowledgment}
\fi

This work is supported by the National Science and Technology Council (NSTC) of Taiwan under grants 112-2628-E-006-012-MY3, 110-2221-E-006-136-MY3, and 112-2634-F-002-006. This work is also supported by the Institute of Information Science (IIS), Academia Sinica, Taiwan.

\ifCLASSOPTIONcaptionsoff
  \newpage
\fi



%
\bibliographystyle{plain}
\bibliography{9-References}
\end{document}